\documentclass[lettersize,journal]{IEEEtran}
\usepackage{amsmath,amsfonts}
\usepackage{array}

\usepackage[caption=false, font=footnotesize, labelformat=empty]{subfig}
\usepackage{textcomp}
\usepackage{stfloats}
\usepackage{url}
\usepackage{verbatim}
\usepackage{graphicx}
\usepackage{cite}
\makeatletter
\def\bstctlcite{\@ifnextchar[{\@bstctlcite}{\@bstctlcite[@auxout]}}
\def\@bstctlcite[#1]#2{\@bsphack
  \@for\@citeb:=#2\do{%
    \edef\@citeb{\expandafter\@firstofone\@citeb}%
    \if@filesw\immediate\write\csname #1\endcsname{\string\citation{\@citeb}}\fi}%
  \@esphack}
\makeatother

\usepackage[ruled,vlined]{algorithm2e}

\usepackage{booktabs}
\usepackage{multirow}

\usepackage[T1]{fontenc}

\usepackage{calc} 

\usepackage[pdfstartview=XYZ,
bookmarks=true,
colorlinks=true,
linkcolor=blue,
urlcolor=blue,
citecolor=blue,
bookmarks=true,
linktocpage=true, 
hyperindex=true
]{hyperref}

\usepackage{orcidlink}

\begin{document}
\bstctlcite{IEEEexample:BSTcontrol} 

\title{Chronology of Multi-Agent Interactions for Provenance of Evolving Information}

\author{Ching-Chun Chang and Isao Echizen

\thanks{C.-C. Chang and I. Echizen are with the Information and Society Research Division, National Institute of Informatics, Tokyo, Japan. I. Echizen is also with the Graduate School of Information Science and Technology, University of Tokyo, and the School of Multidisciplinary Sciences, Graduate University for Advanced Studies (SOKENDAI), Tokyo, Japan.
}
\thanks{Correspondence: C.-C. Chang (email: ccchang@nii.ac.jp)
}
}

\maketitle

\begin{abstract}

Provenance is the chronological history of things, resonating with the fundamental pursuit to uncover origins, trace connections, and situate entities within the flow of space and time. As artificial intelligence advances towards autonomous agents capable of interactive collaboration on complex tasks, the provenance of generated content becomes entangled in the interplay of collective creation, where contributions are continuously revised, extended or overwritten. In a multi-agent generative chain, content undergoes successive transformations, often leaving little, if any, trace of prior contributions. In this study, we investigate the problem of tracking multi-agent provenance across the temporal dimension of generation. We propose a chronological system for post hoc attribution of generative history from content alone, without reliance on internal memory states or external meta-information. At its core lies the notion of symbolic chronicles, representing signed and time-stamped records, in a form analogous to the chain of custody in forensic science. The system operates through a feedback loop, whereby each generative timestep updates the chronicle of prior interactions and synchronises it with the synthetic content in the very act of generation. This research seeks to develop an accountable form of collaborative artificial intelligence within evolving cyber ecosystems.

\end{abstract}

\section{Introduction}

\IEEEPARstart{A}{rtificial} intelligence (AI) emerges from the quest to mimic human minds through the creation of computational machinery that learns through experience and evolves beyond programmed instructions~\cite{Turing:1950aa}. These learning machines extract meaningful representations from data, adapt their behaviours based on feedback, and ultimately develop agency for autonomous interaction with environments~\cite{Hopfield:1982aa, 10.1145/1968.1972, Ackley:1985aa, Rumelhart:1986aa, Sutton:1988aa, 6796673, Mnih:2015aa, LeCun:2015aa, Hafner:2025aa}. Yet, it remains an open question whether \emph{general-purpose intelligence} can be realised within a solitary monolithic neural network~\cite{Fodor:1983aa}. While the \emph{neural scaling law} suggests continuous improvements in generalisability as computational resources and corpora of knowledge increase, an ever-scaling model may still encounter inherent limitations when confronting tasks of sufficient complexity. It may struggle to decompose problems hierarchically, integrate knowledge across diverse domains and sensory modalities, sustain multiple concurrent lines of reasoning, or preserve coherence over extended inferential chains.

\begin{figure*}[t!]
\centering
\includegraphics[width=0.7\linewidth]{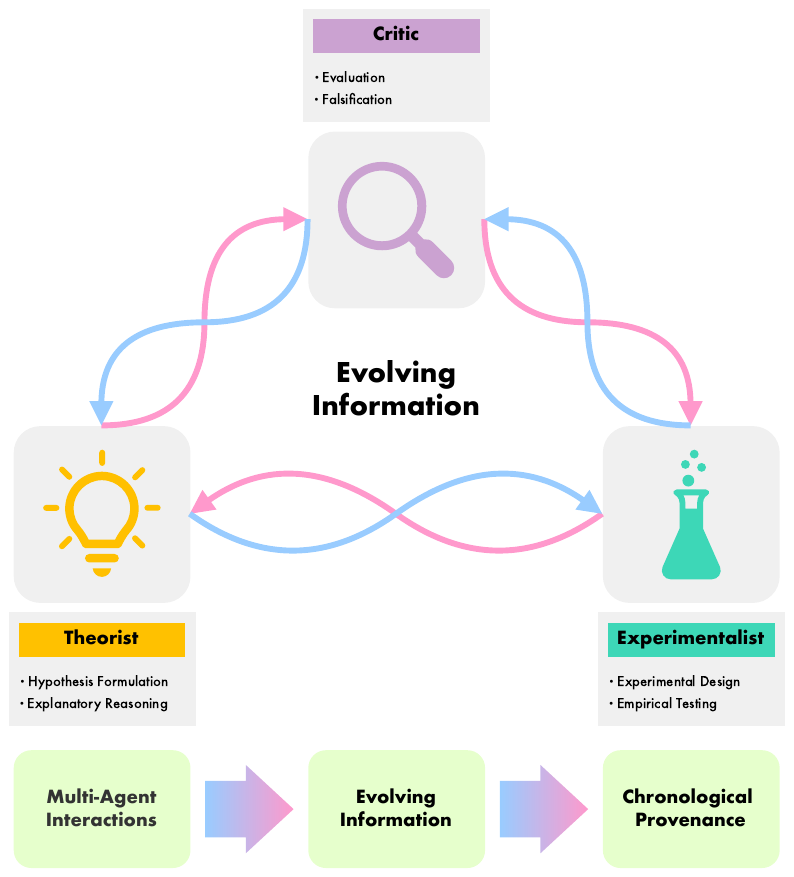}
\caption{A multi-agent system for scientific research, in which information evolves through interactions among specialised agents (theorist, experimentalist and critic), while its chronological provenance remains traceable.}
\label{fig:intro}
\end{figure*}

This calls for an account of \emph{multi-agent collaboration}, in which each agent may possess specialised expertise, and through communication and coordination, these autonomous entities collectively tackle multifaceted tasks~\cite{10.5555/3091574.3091594, Grosz:1996aa, 10.5555/295240.295800, 10.5555/945365.964288, Panait:2005aa, 10.5555/3295222.3295385, NEURIPS2024_ee71a4b1}. Against this backdrop, a fundamental question arises: as multiple agents contribute to the formation of a shared creation, how might one trace the individual contributions of each agent across the temporal dimension of generation? As an illustrative example, consider a scenario of multi-agent collaborative scientific research, in which specialised agents (e.g. a theorist, an experimentalist and a critic) interact autonomously, as shown in Figure~\ref{fig:intro}. Through these interactions, information evolves and yet the final conclusion may leave no trace of whether its validity has been theoretically reasoned, empirically tested, or critically challenged. Without traceability, the interactions amongst multiple agents render the accountability, transparency and trustworthiness of AI fundamentally uncertain.

The concept of \emph{provenance} resonates deeply with the fundamental human pursuit of understanding where we come from and how things came to be, addressing the primordial desire to know origins and connections across space and time. The quest for provenance is a reflection of our intrinsic curiosity about existence and continuity, manifesting across disciplines: astrophysics seeking the origins of the universe; archaeology unearthing ancient civilisations; and genetics reconstructing the evolutionary tree of life. By examining information encoded in genomes, geneticists can infer evolutionary histories, tracing the lineages of organisms across time. Likewise, when analysing a piece of writing, linguists may identify authorial traits through stylistic patterns, at times revealing signs of collaborative composition. In the realm of generative AI, however, these principles and practices face a fundamental challenge. Unlike genomic evidences left by biological evolution or linguistic clues of human authorship, the outputs of multi-agent systems may undergo complete transformation at each step. A subsequent agent may overwrite the content produced by its predecessors to maintain consistency in narrative flow, or continue the task while discarding earlier material, leaving no discernible trace of prior contributions.

One might attempt to adapt the practice of \emph{chain of custody} from forensic science\textemdash chronological documentation recording the seizure, control, transfer and disposition of criminological evidence. In theory, such an external log could record which agent acted at each timestep, preserving a complete history of the generative process. Yet this forensic practice remains inherently fragile. External logs, or metadata, can become detached from the content it describes when transferred across platforms or corrupted during transmission, undermining the integrity of the provenance it was meant to secure. In the absence of metadata, how might provenance be preserved in a form analogous to a chain of custody, maintaining a signed and time-stamped record at each transaction amongst collaborating AI agents?
 
 In this study, we introduce a chronological system for tracking provenance in the context of multi-agent collaboration. At its core is the concept of chronicles\textemdash symbolic sequences that represent the chronologically ordered identities of agents throughout the generative process. The system operates through a feedback loop, in which each generative step updates the chronicle of prior interactions and synchronises it with the generated content in a steganographic manner. The scope of this study centres on generative chains of natural language created by foundation AI models. Each valid chronicle is associated with a binary codeword that defines a lexical subset of the vocabulary. The present state of the chronicle is then embedded into the generated text via a biased language generation process conditioned on the associated subset. The chronicle can subsequently be retrieved from the text through statistical analysis. This chronological system archives multi-agent provenance within the text itself, evolving alongside the act of generation without reliance on external meta-information.

\section{Provenance}
This section reviews foundational methodologies for attributing the provenance of digital content. We examine three primary techniques, including \emph{metadata annotation}, \emph{fingerprinting}, and \emph{watermarking}, outlining their principal capabilities and inherent limitations.

\subsection{Metadata Annotation}
Metadata annotation involves attaching descriptive information to data, detailing aspects such as origin of data, identity of author, time of creation, and history of usage. This archival practice facilitates data organisation and retrieval, forming the backbone of many provenance management systems~\cite{267415, 10.5555/1267359.1267363, 10.1145/1629080.1629082}. Cryptographic techniques are often applied for certifying signatures~\cite{10.1145/359340.359342}, timestamps~\cite{Haber:1991aa} and audit trails~\cite{10.1145/317087.317089}. However, the fundamental limitation of metadata lies in its extrinsic nature. Metadata exists apart from the content it describes, making it susceptible to removal, manipulation or disassociation. This separation can compromise traceability, particularly in environments where imperfect transmission, format conversion or adversarial actions may occur.

\subsection{Fingerprinting}
Fingerprinting refers to the process of generating identifiable representations, known as fingerprints, that can uniquely identify objects~\cite{705568, 841169, 1188750}. Fingerprinting methods can be broadly classified into two primary categories: \emph{cryptographic hashing} and \emph{perceptual hashing}. Cryptographic hashing computes fixed-size digests from arbitrary-length inputs using one-way hash functions designed to be sensitive to input changes with low probability of collisions~\cite{Damgard:1989aa, 10.5555/118209.118249, 10.1007/3-540-68697-5_1}. It is characterised by the avalanche effect, whereby a minimal change in the input propagates and results in a drastically different hash output. However, this sensitivity to data integrity, albeit essential for tamper-proofing, poses limitations in scenarios where content-preserving transformations are expected. Perceptual hashing extracts robust content-dependent features from data that remain stable under content-preserving transformations, thereby enabling similarity-based identification~\cite{899541, 1709989, 1634363}. It is particularly applicable to multimedia content, which is often subject to compression, resampling, or format conversion. By determining the degree of similarity between two pieces of content, it facilitates fuzzy matching for applications such as duplicate detection and similarity search. However, this robustness to modifications comes at the cost of reduced discriminability, potentially causing collisions between different contents that share similar global structures. Cryptographic hashing offers high sensitivity for discriminability but fails under content transformations, whereas perceptual hashing provides robustness against content transformations but lacks strong guarantees of uniqueness. This trade-off between collision resistance and fault tolerance reflects a fundamental dilemma between sensitivity and robustness.

\subsection{Watermarking}
Watermarking is the practice of embedding auxiliary information into the content subject to imperceptibility constraints with respect to human perception, thereby enabling self-contained traceability without reliance on external metadata~\cite{413536, 650120, 687830, 771066, 771072, 771068, 959098}. A watermark can carry provenance-related information such as unique identifiers and timestamps, providing the capability of collision resistance. In addition, it can be embedded in a way that survives common content-preserving operations such as compression, photometric distortion or geometric transformation, thus offering robust proof of provenance in multimedia distribution. The concept of watermarking has been applied to generative foundation models for provenance tracing of AI-generated synthetic content~\cite{Dathathri:2024aa}. A representative methodology is based on biasing the token sampling process during text generation, controlling the distribution of selected tokens to form statistically detectable patterns~\cite{pmlr-v202-kirchenbauer23a}. It follows the principle of \emph{zero-bit watermarking}, in which no explicit information is embedded; rather, provenance is verified by testing for the presence or absence of a given watermark. In multi-agent environments, however, zero-bit watermarking may not be directly applicable because it typically attributes each piece of content to a single source authority. While in principle it is possible for multiple watermarks to coexist within a single object without mutual interference or overwriting each other, such coexistence is not always guaranteed~\cite{petrov2025on}. Even when multiple watermarks coexist, the temporal order of sequential embedding may not be reliably determined. This uncertainty calls for methodological reconsideration under circumstances where multiple agents interact and collaborate in a generative chain.

\section{Chronology}
This section introduces the proposed chronological system for tracking multi-agent provenance in language generation. We formalise the notion of a chronicle, present a scalable codebook construction, describe the feedback loop for recursively updating chronicles, and detail the encoding and decoding procedures that enable post hoc recovery of generative histories from text alone.

\begin{figure}[t!]
\centering
\includegraphics[width=0.99\linewidth]{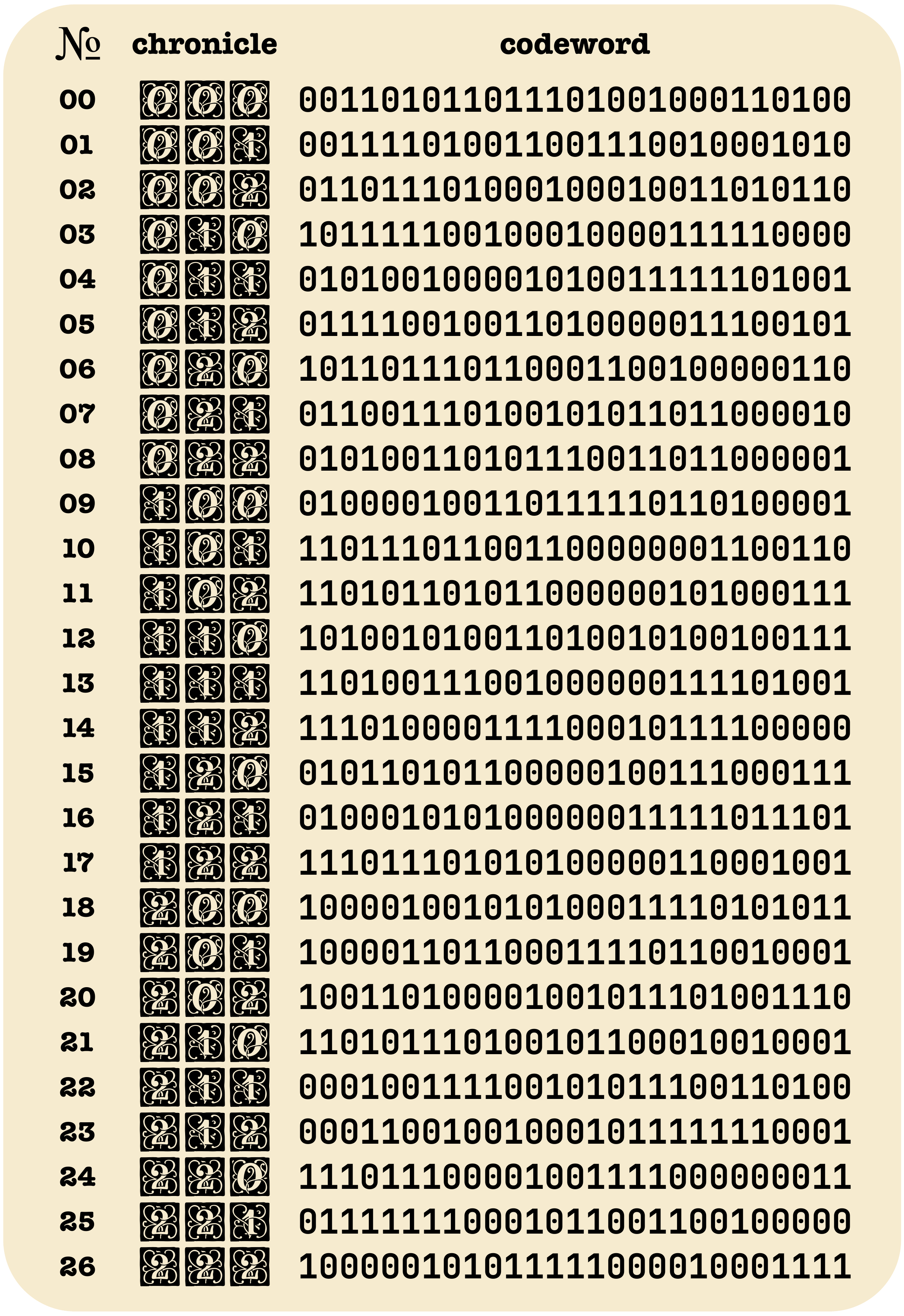}
\caption{An example of the base codebook for an agent population 2 and chronicle length 3 with the vocabulary coverage rate set to 0.5, where the codewords are shown prior to expansion to the full vocabulary size.}
\label{fig:codebook}
\end{figure}

\subsection{Chronicle Definition}

Consider a set of $n$ distinct AI agents powered by foundation models. A \emph{chronicle} is defined as a symbolic chain of length $T$ (i.e. the number of generative steps), represented by
\begin{equation}
    \boldsymbol{x} = [x_1, x_2, \dots, x_T],
\end{equation}
where each symbol $x_t \in \{0, 1, \dots, n\}$ indicates the identity of the agent assigned at timestep $t$, and the symbol $0$ denotes a null or unassigned agent. The set of all valid chronicles is defined as
\begin{equation}
    \mathcal{X} = \{0, 1, \dots, n\}^T.
\end{equation}
Each unique chronicle $\boldsymbol{x}$ is associated with a corresponding binary \emph{codeword} 
\begin{equation} 
	\boldsymbol{c}({\boldsymbol{x}} )\in \{0,1\}^{|\mathcal{V}|}, 
\end{equation} 
where $\mathcal{V}$ denotes the vocabulary of the underlying foundation model. The codeword marks a sparse subset of the vocabulary (e.g. with 50\% of entries set to 1). The tokens corresponding to indices marked with $1$ in the codeword are biased towards during generation. The collection of all codewords defines the \emph{codebook} 
\begin{equation} 
	\mathcal{C} = \{ \boldsymbol{c}({\boldsymbol{x}}) \mid \boldsymbol{x} \in \mathcal{X} \}. 
\end{equation}
The cardinality of the codebook is $ |\mathcal{X}| = (n+1)^T $, encompassing all possible chronicle configurations.

\begin{figure*}[t!]
\centering
\includegraphics[width=0.99\linewidth]{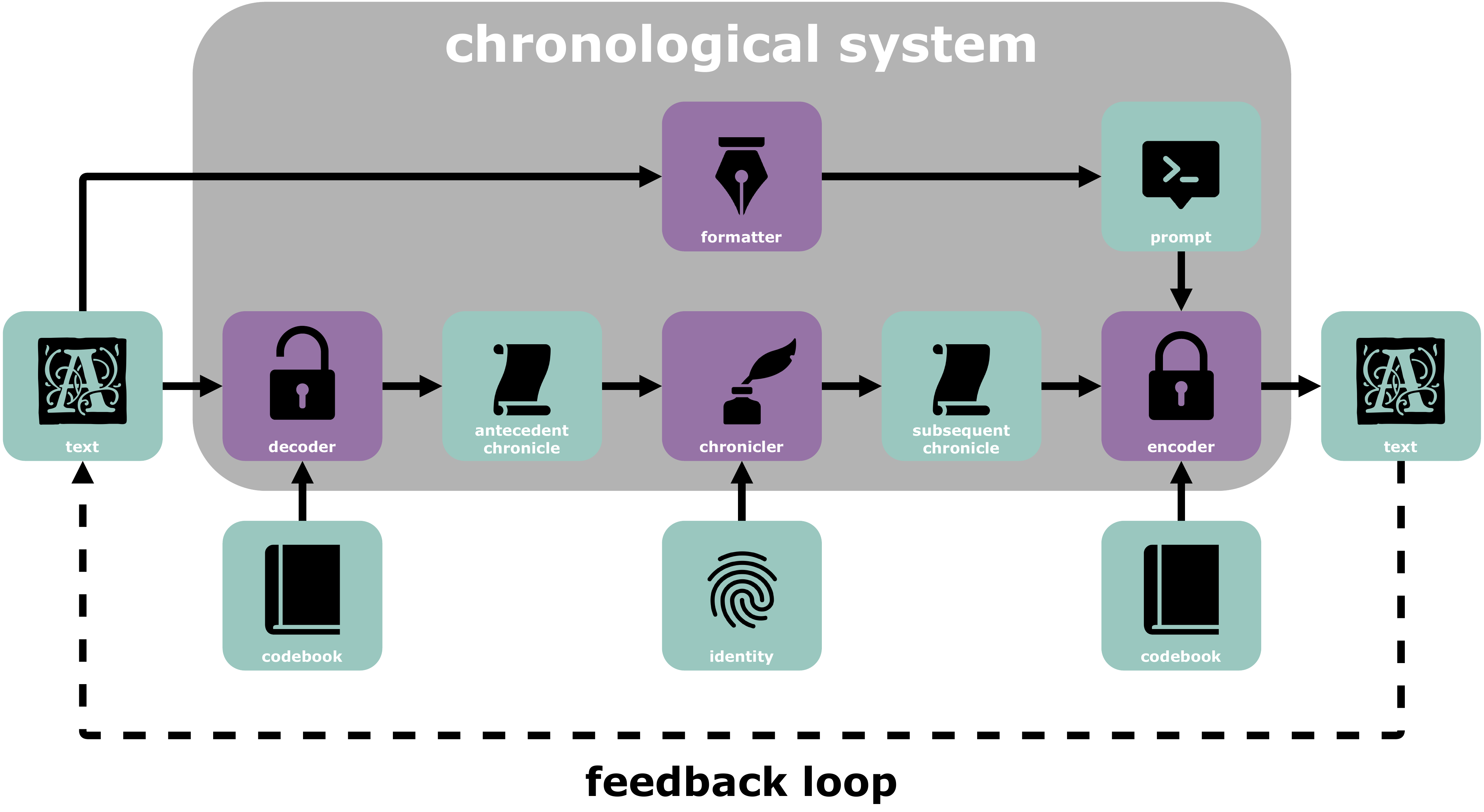}
\caption{Overview of the chronological system for provenance tracking through encoding, decoding and updating of chronicles within a feedback loop.}
\label{fig:overview}
\end{figure*}

\subsection{Codebook Construction}
A codebook $\mathcal{C}$ is of dimension $|\mathcal{X}| \times |\mathcal{V}|$, where $|\mathcal{X}| = (n+1)^T$ denotes the number of all valid chronicles and $|\mathcal{V}|$ is the vocabulary size of the language model. Direct generation and storage of such a codebook with unique codewords can be computationally prohibitive and memory-intensive, particularly as the agent population, the chronicle length and the vocabulary size scale up. To keep the construction within a tractable combinatorial regime, we formulate a scalable codebook generation strategy that first constructs a base codebook in a reduced space of dimension $d$, resulting in a matrix of size $|\mathcal{X}| \times d$. Each base codeword is then expanded to match the full vocabulary size $|\mathcal{V}|$ through repetition and structured padding. In the base codebook, each codeword is generated randomly as a binary vector of constant Hamming weight $\lfloor \rho \cdot d \rfloor$, where $\rho \in (0,1)$ is a vocabulary coverage rate. A generated vector is added to the codebook only if its pattern does not duplicate any previously stored codeword, and this stochastic generation process repeats until $|\mathcal{X}|$ unique codewords are obtained. Each base codeword is then repeated as many times as possible to reach length $|\mathcal{V}|$, with the remaining positions padded to satisfy the target Hamming weight $\lfloor \rho \cdot |\mathcal{V}| \rfloor$. To sum up, the codebook is constructed by first generating $|\mathcal{X}|$ unique base codewords of length $d$ with weight $\lfloor \rho \cdot d \rfloor$ and then expand each base codeword to length $|\mathcal{V}|$ by repetition and padding to achieve the target weight $\lfloor \rho \cdot |\mathcal{V}| \rfloor$, as shown in Figure~\ref{fig:codebook}. In practice, we set $\rho = 0.5$ so that a sufficiently large subset of the vocabulary is associated with higher sampling likelihood during text generation. We further set $d = |\mathcal{X}|$, under which the size of the constant-weight code space $\binom{d}{\lfloor \rho d \rfloor}$ grows combinatorially and is substantially larger than $|\mathcal{X}|$, thereby ensuring that sufficiently many distinct base codewords exist for random sampling.

\subsection{Chronological System with Feedback Loop}
At each timestep $t$, an \emph{antecedent chronicle} $\boldsymbol{x}^{(t-1)}$ is decoded and then updated into a \emph{subsequent chronicle} $\boldsymbol{x}^{(t)}$ to be encoded, as illustrated in Figure~\ref{fig:overview} and presented in Algorithm~\ref{alg:system}. The process begins with an initial chronicle set to the all-zero sequence. The chronicler updates the antecedent chronicle by inserting the informed agent identity at position $t$, leaving the remaining entries as zeros. In other words, the subsequent chronicle is formed by concatenating a prefix of agent identities up to the preceding timestep, an infix corresponding to the current agent identity, and a suffix of zero symbols to match the predefined chronicle length, yielding 
\begin{equation}
\boldsymbol{x}^{(t)} = \boldsymbol{x}^{(t-1)}_{1:t-1} \mathbin{\Vert} x_t \mathbin{\Vert} \boldsymbol{0}_{t+1:T} .
\end{equation}
The up-to-date chronicle is passed to the encoder, which embeds it into the text during the generation process of the designated agent. Once the text is generated, it is returned to the decoder, which retrieves the embedded chronicle. For the next generation step, the chronicle is updated, and the generated text is formatted into a prompt that includes the task description and, optionally, a persona that specifies the role-specific traits, behavioural constraints or stylistic preferences. The continuation of this feedback loop recursively updates the chronicle across timesteps, thereby enabling the tracking of agentic provenance throughout the generative process. Note that the update of the chronicle may optionally be performed without explicit knowledge of the current timestep; instead, the position of the next zero symbol serves as a clue for inferring it. While the relaxation of timestep awareness offers flexibility in application scenarios where timestep tracking is unavailable or undesirable, it sacrifices fault tolerance. The update mechanism under this relaxation becomes sensitive to errors in preceding steps, where a single erroneous chronicle may trigger a cascading collapse across the remaining timesteps.

\begin{algorithm}[t!]
\caption{Chronological System}
\label{alg:system}

\vspace{1ex}
\tcp{----- functions -----}
\SetKwFunction{FEecoder}{Encoder}
\SetKwProg{Fn}{Function}{:}{}
\Fn{\FEecoder{\textnormal{$\textit{\textbf{prompt}}$, $\mathcal{C}$, $\boldsymbol{x}$}}}{
    retrieve codeword $\boldsymbol{c}(\boldsymbol{x})$ from codebook $\mathcal{C}$\\
    
    \While{$\textnormal{terminal criterion not met}$}{
    	predict logits $\ell_v$ conditioned on $\textit{\textbf{prompt}}$\\
    	apply bias according to codeword $\boldsymbol{c}(\boldsymbol{x})$:\\
    	\ForEach{$v \in \mathcal{V}$}{
        	\eIf{$c_v = 1$}{
            	$\tilde{\ell}_v \gets \ell_v + \delta$\;
        	}{
            	$\tilde{\ell}_v \gets \ell_v$\;
        	}
    	}
    	convert logits to probabilities $p_v = \operatorname{softmax}(\tilde{\ell}_v)$\\
    	sample a token from probability distribution $p_v$\\
	}
	detokenise tokens $\boldsymbol{v}$ to $\textit{\textbf{text}}$\\
	\KwRet{$\textit{\textbf{text}}$}\\
}

\vspace{1ex}
\SetKwFunction{FDecoder}{Decoder}
\SetKwProg{Fn}{Function}{:}{}
\Fn{\FDecoder{\textnormal{$\textit{\textbf{text}}$, $\mathcal{C}$}}}{
    tokenise $\textit{\textbf{text}}$ into tokens $\boldsymbol{v}$\\
    generate all valid chronicles $\mathcal{X}$\\
    \ForEach{$\boldsymbol{x} \in \mathcal{X}$}{
        retrieve codeword $\boldsymbol{c}(\boldsymbol{x})$ from codebook $\mathcal{C}$\\
        compute frequency $f(\boldsymbol{x}) = \sum_{i=1}^{|\boldsymbol{v}|} \mathbb{I}(c_{v_i} = 1)$\\
    }
    infer likeliest chronicle $\hat{\boldsymbol{x}} = \arg\max_{\boldsymbol{x}} f(\boldsymbol{x})$\\
    \KwRet{$\hat{\boldsymbol{x}}$}\\
}

\vspace{1ex}
\tcp{----- main program -----}
set vocabulary $\mathcal{V}$\\
set number of agents $n$\\
set number of timesteps $T$\\
initialise codebook $\mathcal{C} \gets \{ \boldsymbol{c}({\boldsymbol{x}}) \mid \boldsymbol{x} \in \mathcal{X} \}$\\where $\mathcal{X} = \{0, 1, \dots, n\}^T$ and $\boldsymbol{c}({\boldsymbol{x}} )\in \{0,1\}^{|\mathcal{V}|}$\\
initialise chronicle $\boldsymbol{x}^{(0)} \gets \boldsymbol{0}_{1:T}$\\

\For{$t \gets 1$ \KwTo $T$}{
	assign agent identity $x_t$ for generation process\\
	construct $\textit{\textbf{prompt}}$ for generation process\\
	update chronicle:\\
	$\boldsymbol{x}^{(t)} \gets \boldsymbol{x}^{(t-1)}_{1:t-1} \Vert x_t \Vert \boldsymbol{0}_{t+1:T}$\\
	encode chronicle:\\
	$\textit{\textbf{text}} \gets$ \FEecoder{\textnormal{$\textit{\textbf{prompt}}$, $\mathcal{C}$, $\boldsymbol{x}^{(t)}$}}\\
	
	decode chronicle:\\
	$\boldsymbol{x}^{(t)} \gets$ \FDecoder{\textnormal{$\textit{\textbf{text}}$, $\mathcal{C}$}}\\
}
\end{algorithm}

\subsection{Chronicle Encoder}
\begin{figure*}[t!]
\centering
\includegraphics[width=0.95\linewidth]{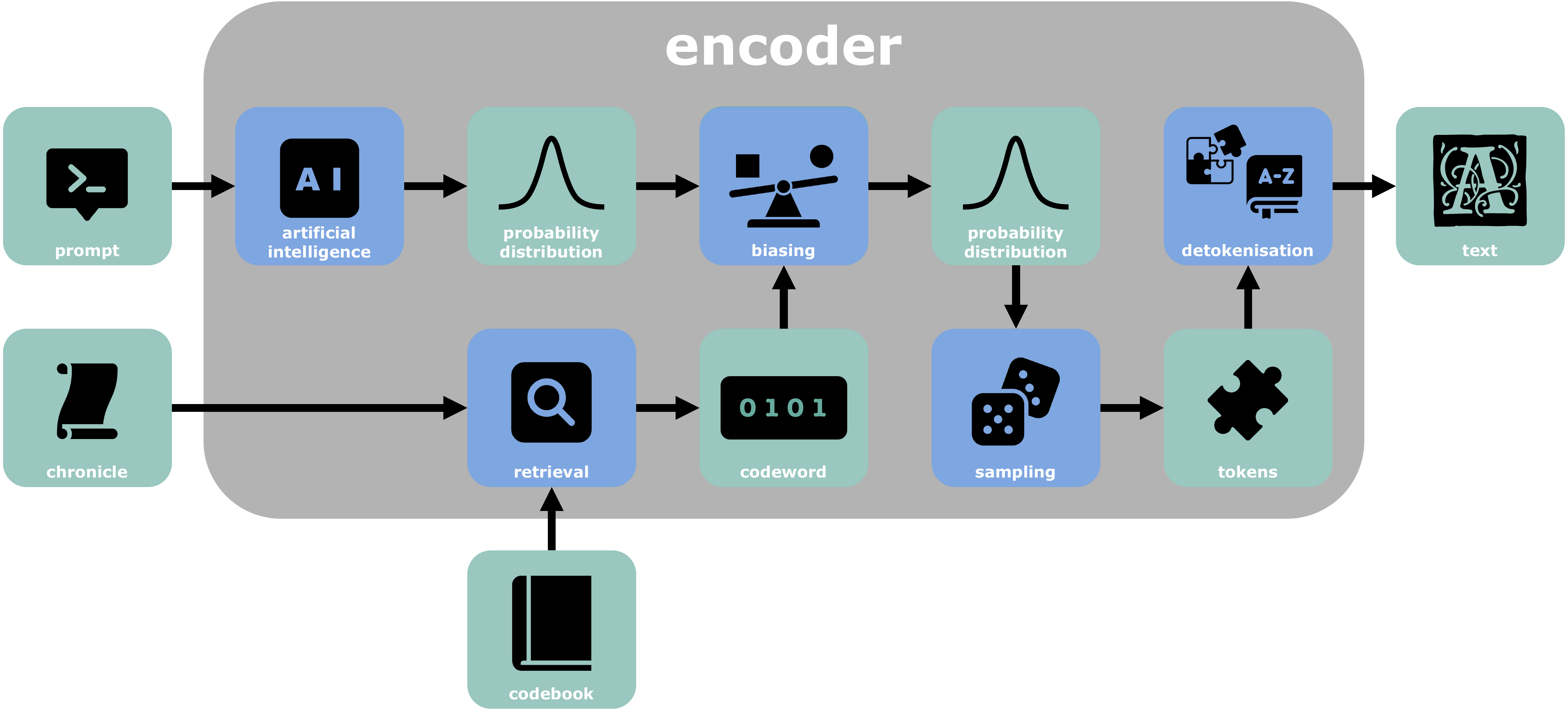}
\caption{Procedure of chronicle encoding, where the chronicle is embedded into the generated text through biased token sampling during language generation.}
\label{fig:encoding}
\end{figure*}

At each timestep $t$, the encoder receives a prompt and an up-to-date chronicle $\boldsymbol{x}^{(t)}$, along with access to a predefined codebook, as illustrated in Figure~\ref{fig:encoding}. The codeword corresponding to the given chronicle $ \boldsymbol{c}({\boldsymbol{x}^{(t)}})$ is retrieved from the codebook. Given the prompt, the designated agent predicts an unnormalised probability distribution (i.e. logits) over the entire vocabulary. Let $\ell_v$ denote the logit associated with token $v \in \mathcal{V}$, and let $c_v$ denote the corresponding binary digit in the codeword. Each logit is then selectively biased to favour tokens marked in the codeword; that is,
\begin{equation}
    \tilde{\ell}_v =
    \begin{cases}
        \ell_v + \delta, & \text{if } c_v = 1, \\
        \ell_v, & \text{otherwise},
    \end{cases}
\end{equation}
where $\delta > 0$ is a bias strength parameter. The biased logits are then normalised via the softmax function to obtain a probability distribution
\begin{equation}
	p_v = \operatorname{softmax}(\tilde{\ell}_{v}) = \frac {e^{\tilde{\ell}_{v}}} {\sum _{i=1}^{ |V| }e^{\tilde{\ell}_{i}}}.
\end{equation}
A token is sampled from the probability distribution, and the text generation process continues until an end criterion is met, such as reaching a maximum token limit or a terminal state. The resulting sequence of tokens $\boldsymbol{v}$ is then detokenised to form natural language text, within which the chronicle is implicitly embedded via lexical bias.

\subsection{Chronicle Decoder}
\begin{figure*}[t!]
\centering
\includegraphics[width=0.95\linewidth]{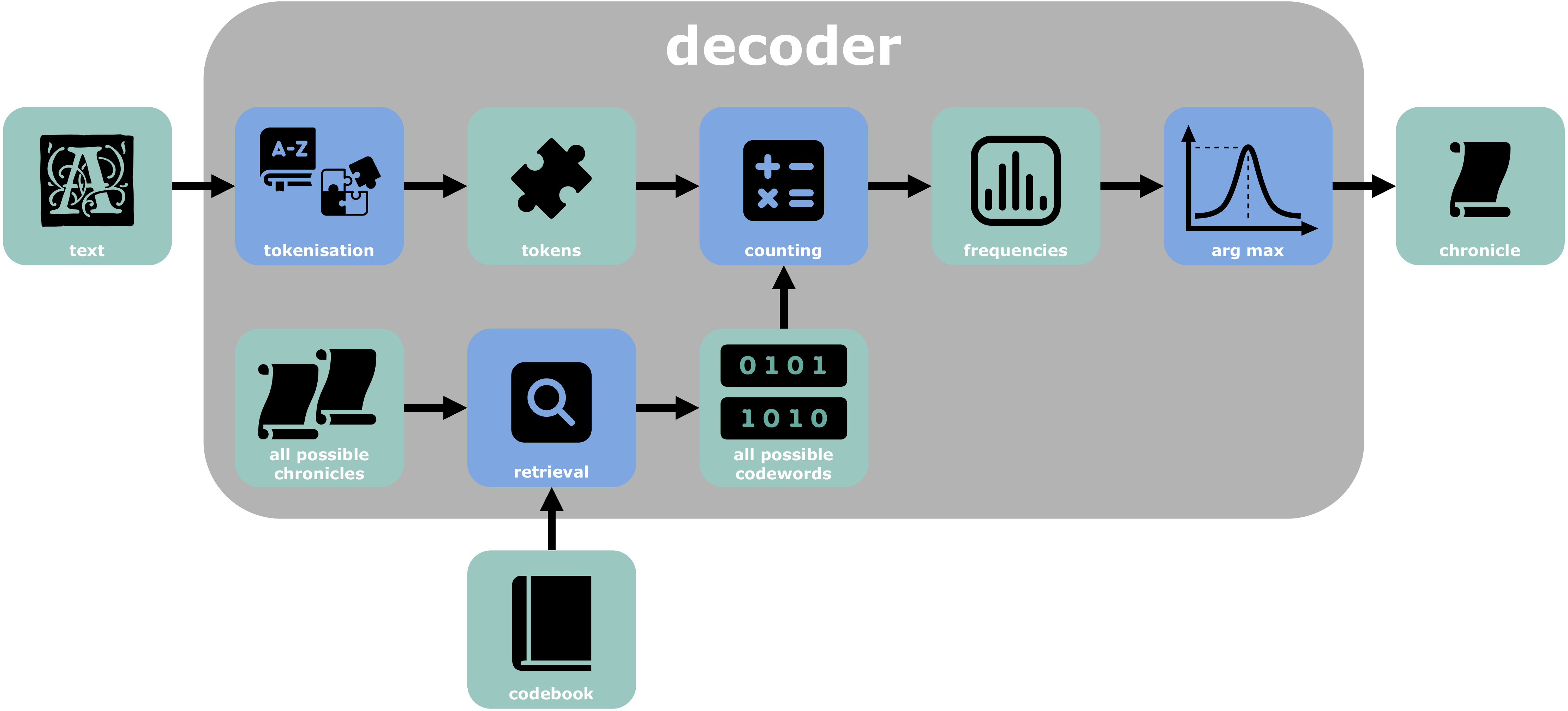}
\caption{Procedure of chronicle decoding, where the chronicle is retrieved from the generated text through statistical analysis of lexical choices.}
\label{fig:decoding}
\end{figure*}

Given the generated text, the decoder estimates the embedded chronicle by comparing the lexical statistics computed over all possible codewords, as illustrated in Figure~\ref{fig:decoding}. The generated text is first tokenised into a sequence of tokens $\boldsymbol{v}$. For each possible chronicle $\boldsymbol{x}$, the corresponding codeword $ \boldsymbol{c}(\boldsymbol{x}) $ is retrieved from the codebook. The frequency of matches $f(\boldsymbol{x})$ is then computed based on the number of marked tokens in the codeword that appear in the generated text; that is,
\begin{equation}
    f(\boldsymbol{x}) = \sum_{i=1}^{| \boldsymbol{v} |} \mathbb{I} (c_{v_i} = 1) ,
\end{equation}
where $\mathbb{I}$ denotes the indicator function that returns $1$ if the condition holds and $0$ otherwise. The most probable chronicle $\hat{\boldsymbol{x}}$ is then estimated by selecting the one with the maximal frequency:
\begin{equation}
    \hat{\boldsymbol{x}} = \arg\max_{\boldsymbol{x} \in \mathcal{X}} f(\boldsymbol{x}).
\end{equation}
The decoded chronicle $\hat{\boldsymbol{x}}$ serves as the reconstructed trace of agentic interactions, enabling post hoc forensic attribution from the generated text alone.

\begin{figure}[t!]
\centering
\includegraphics[width=0.9\linewidth]{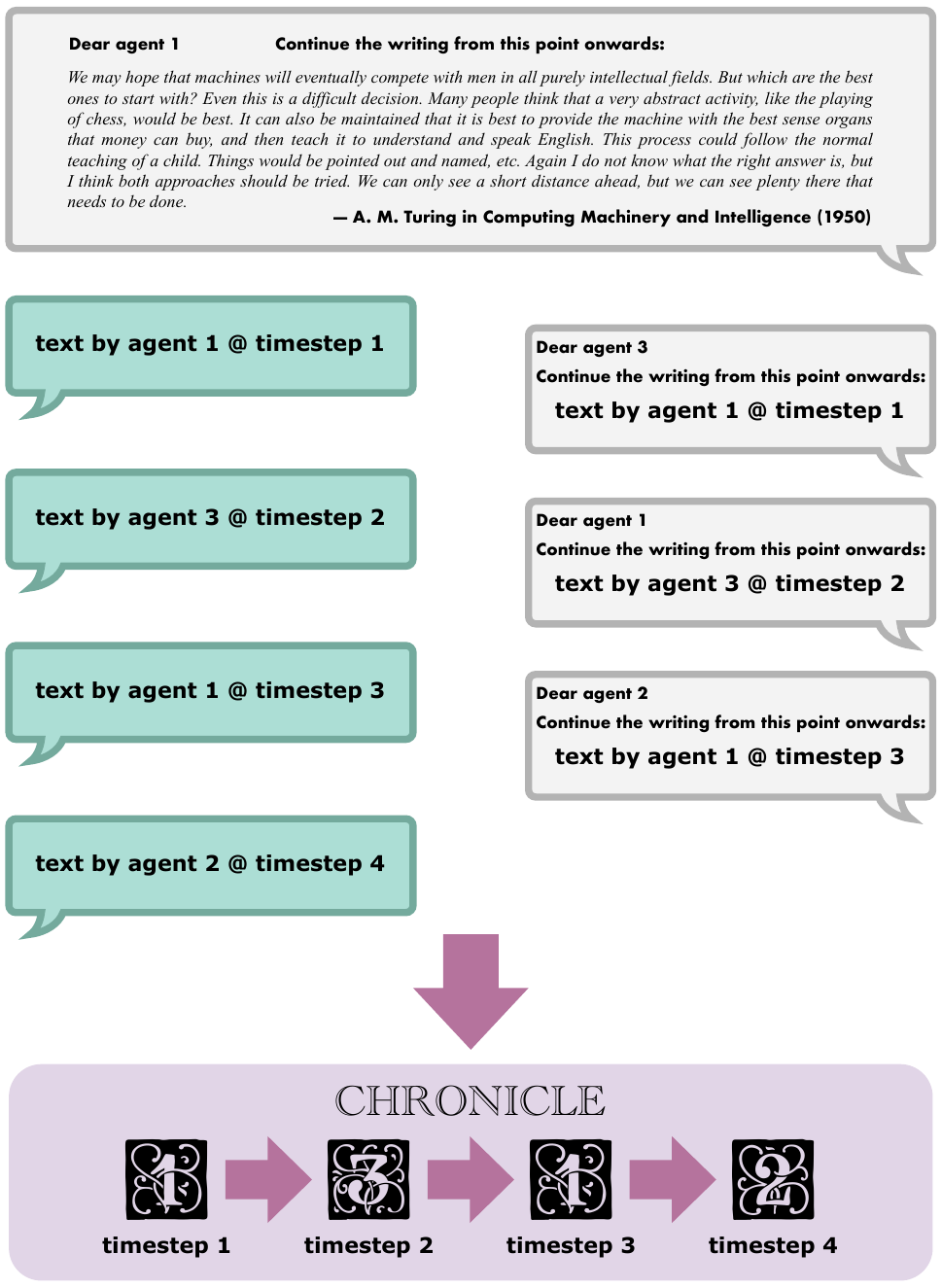}
\caption{Continual generative chain with multiple agents,  where the chronicle is inferred post hoc from the generated content at the final timestep.}
\label{fig:vis_chronicle}
\end{figure}

\section{Evaluation}
This section evaluates the proposed chronological system through simulations of continual language generation involving multiple agents, in which the content is subject to successive transformations. We assess the complexity of the chronicle space, the accuracy of the recovered provenance, the quality of the generated text and the robustness against adversarial attacks under varying experimental conditions.

\subsection{Experimental Setup}

\subsubsection*{Data}
We adopted the final paragraph from \emph{Computing Machinery and Intelligence} by A. M. Turing as the initial seed text. At each generation step, an agent was randomly selected and prompted with the instruction: \emph{Continue the writing from this point onwards}, as illustrated in Figure~\ref{fig:vis_chronicle}. This process forms a \emph{Markovian chain} of generation, where the prompt provided to each agent depends solely on the text generated by its immediate predecessor. This experimental design reflects a continual writing process that operates without memory of the full generation history. As a result, little, if any, trace of past agent assignments is preserved in the text, posing a challenge for recovering the underlying chronology of agent participation.

\subsubsection*{Agents}
All agents were instantiated from the same open-source foundation language model, Llama with 1 billion parameters, chosen for its efficiency and practicality in deployment on lightweight hardware. The generation process was controlled by standard sampling parameters to regulate the trade-off between diversity and coherence. The temperature was set to 0.3 to moderately sharpen the probability distribution over tokens, controlling the degree of randomness in generation. The sampling was further restricted by selecting from the top 1000 most probable tokens (hard-threshold sampling) and from the set of most probable tokens with cumulative probability more than 0.9 (nucleus sampling). The maximum token length was limited to 150 tokens per generation step.

\subsubsection*{Configurations}
The agent population $n$ was varied from 2 to 4, the chronicle length $T$ from 3 to 6, and the bias strength $\delta$ from 1 to 3. For each experimental configuration, we conducted 100 trials with varying random seeds to obtain reliable statistics. Agent identities are randomly assigned at each timestep for each trial without role-specific prompting. As such, agent identity serves solely as a symbolic index in the chronicle, rather than representing specialised functionalities or distinct generative behaviours.

\begin{figure}[t!]
\centering
\includegraphics[width=0.9\linewidth]{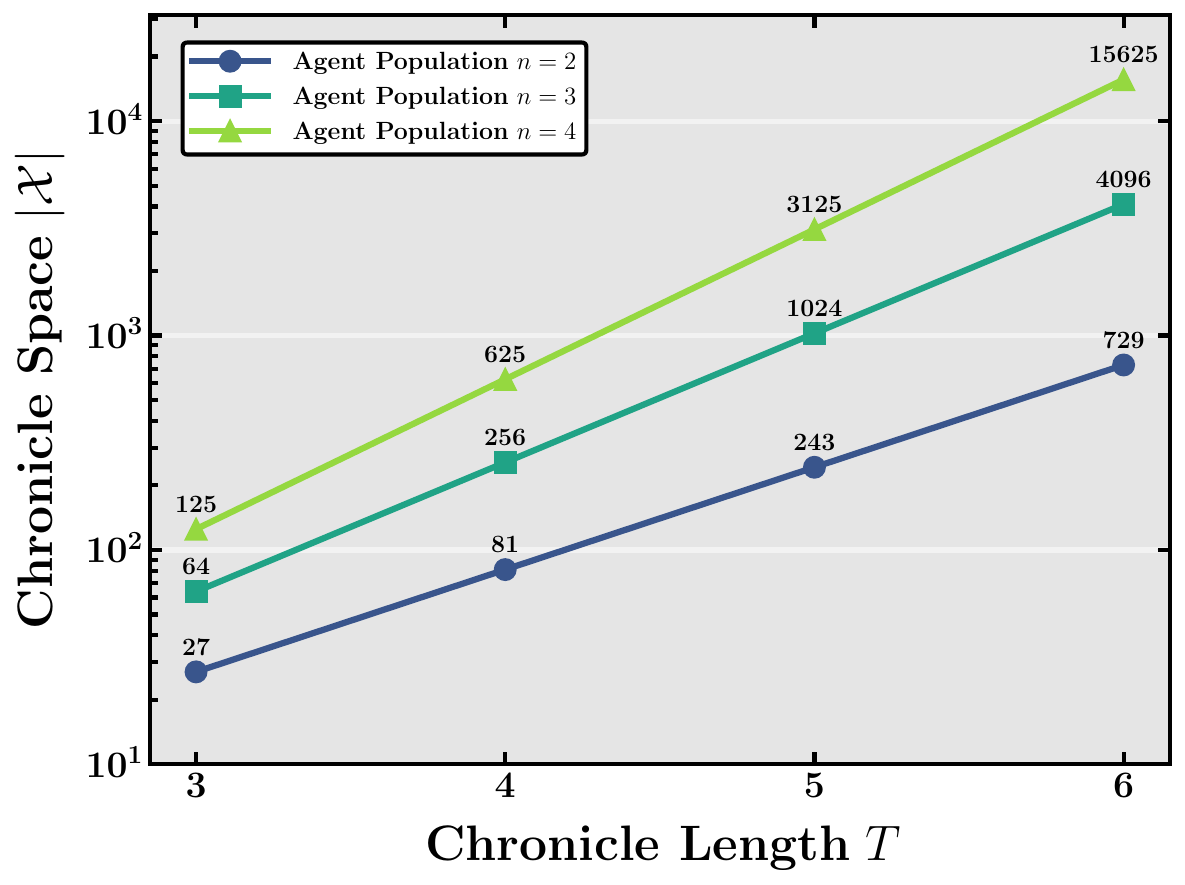}
\caption{Combinatorial scaling of the chronicle space $|\mathcal{X}|$ as a function of chronicle length $T$ and agent population $n$.}
\label{fig:code_complexity}
\end{figure}

\begin{figure*}[t!]
    \centering
    \subfloat[Length 3 \& Population 2]{%
        \includegraphics[width=0.32\linewidth]{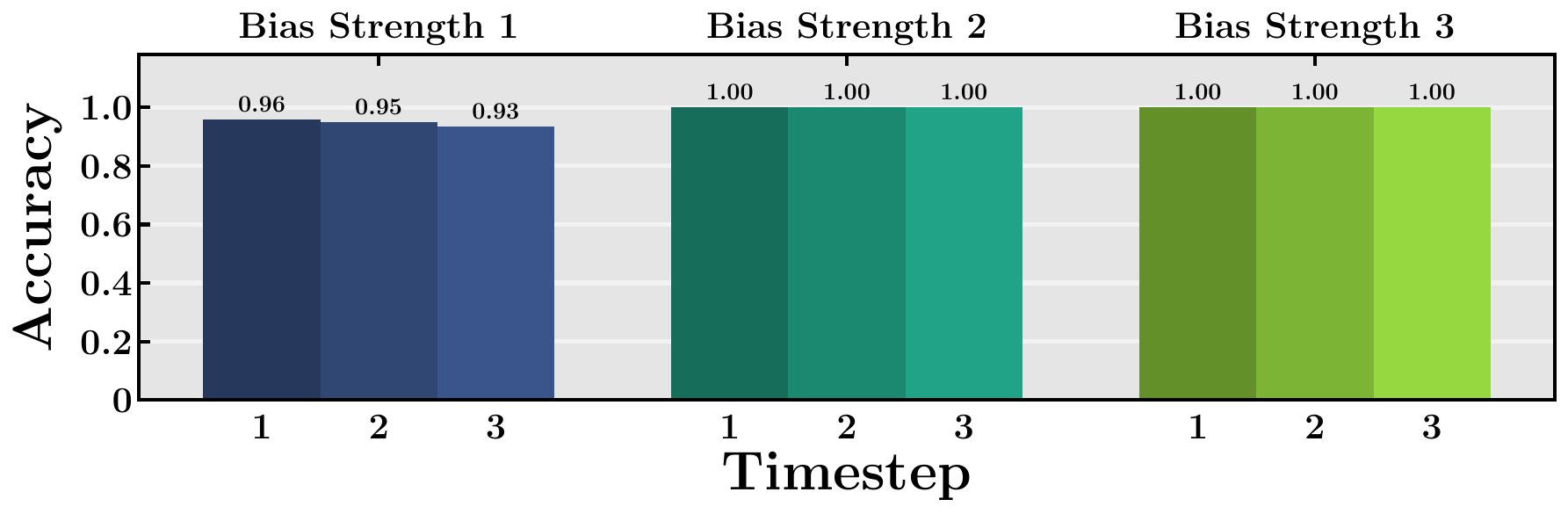}
    }
    \hfill
    \subfloat[Length 3 \& Population 3]{%
        \includegraphics[width=0.32\linewidth]{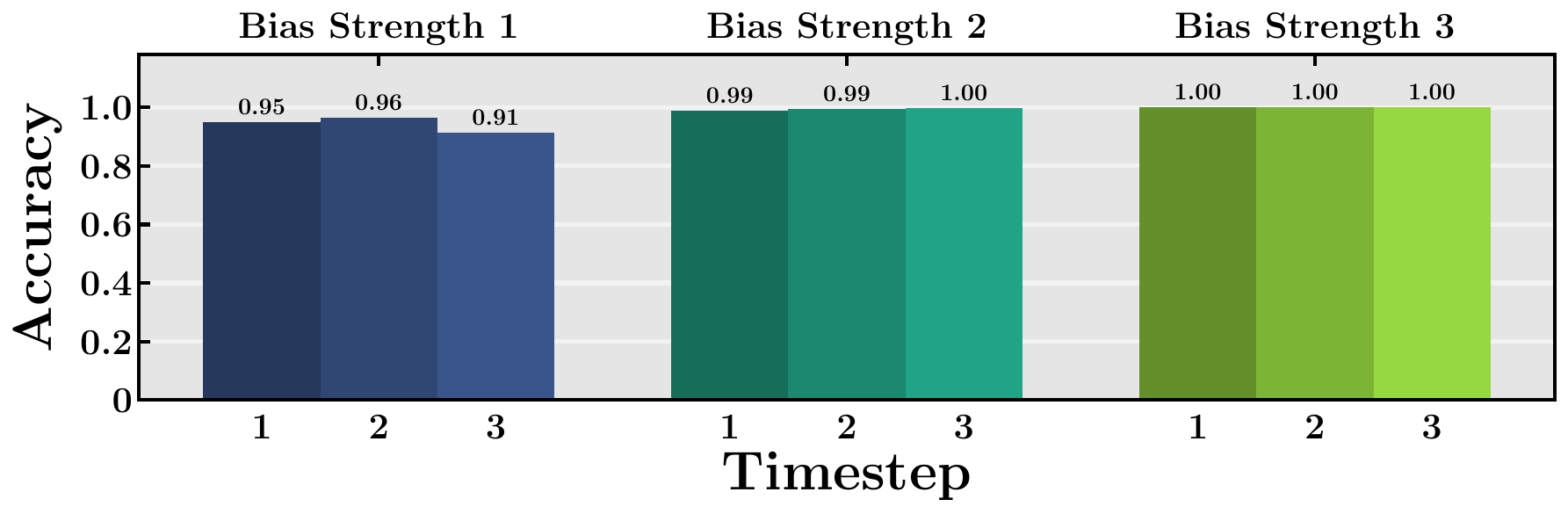}
    }
    \hfill
    \subfloat[Length 3 \& Population 4]{%
        \includegraphics[width=0.32\linewidth]{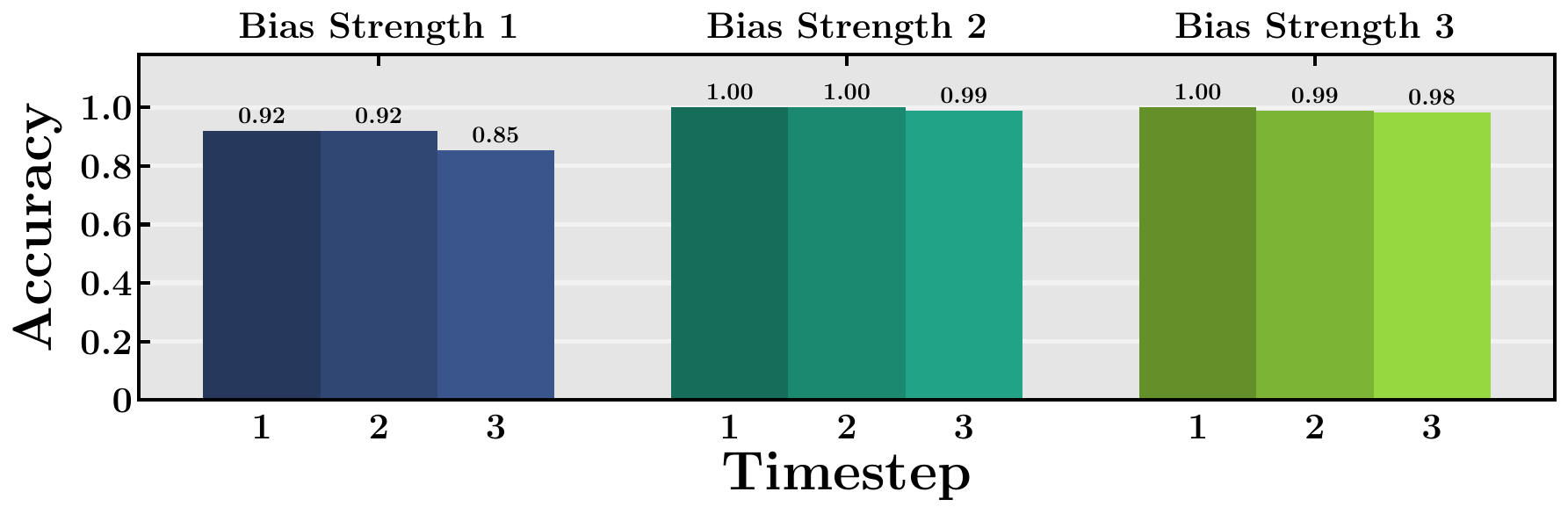}
    }

    \vfill

    \subfloat[Length 4 \& Population 2]{%
        \includegraphics[width=0.32\linewidth]{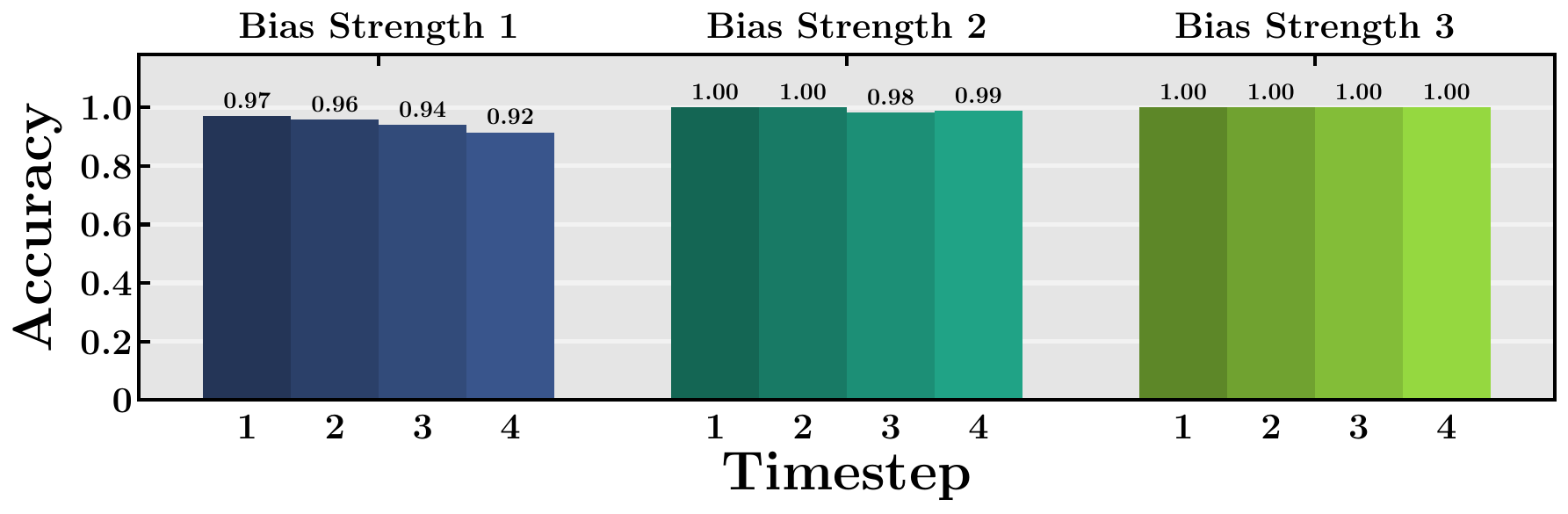}
    }
    \hfill
    \subfloat[Length 4 \& Population 3]{%
        \includegraphics[width=0.32\linewidth]{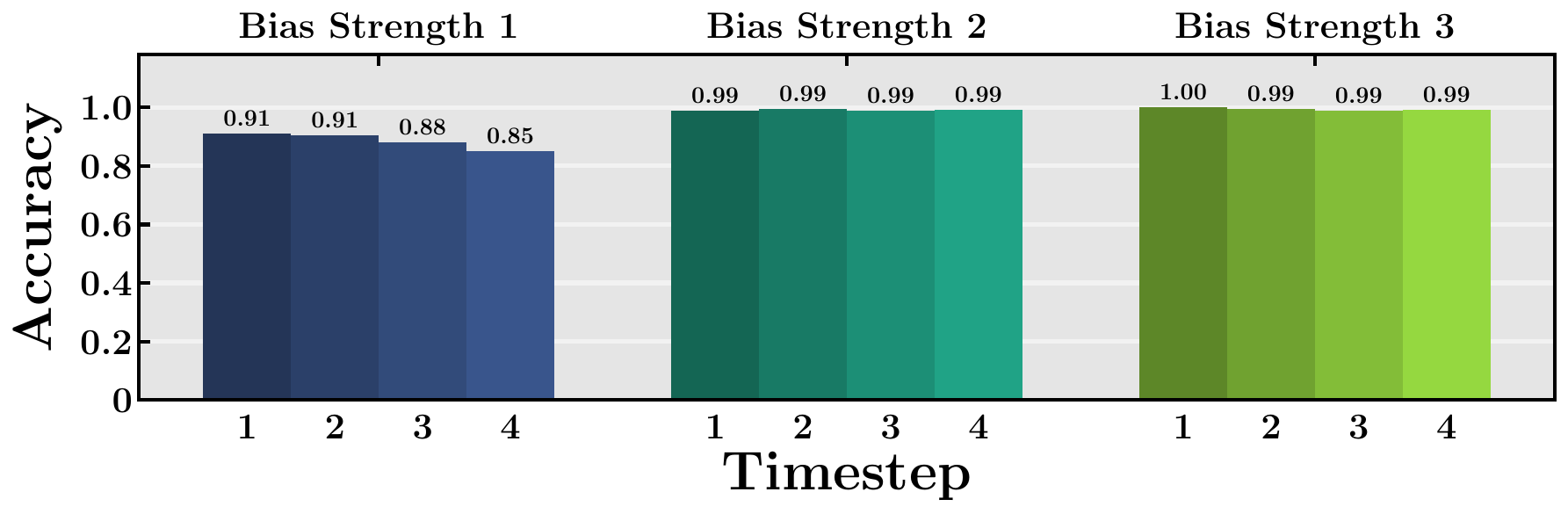}
    }
    \hfill
    \subfloat[Length 4 \& Population 4]{%
        \includegraphics[width=0.32\linewidth]{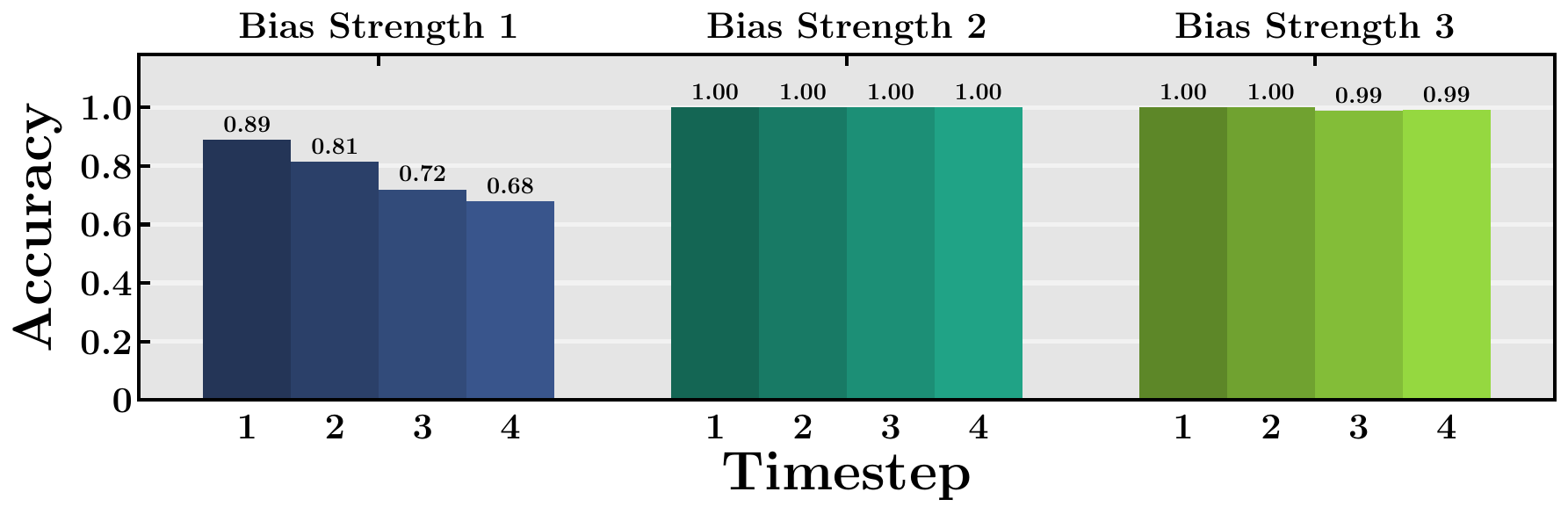}
    }

    \vfill

    \subfloat[Length 5 \& Population 2]{%
        \includegraphics[width=0.32\linewidth]{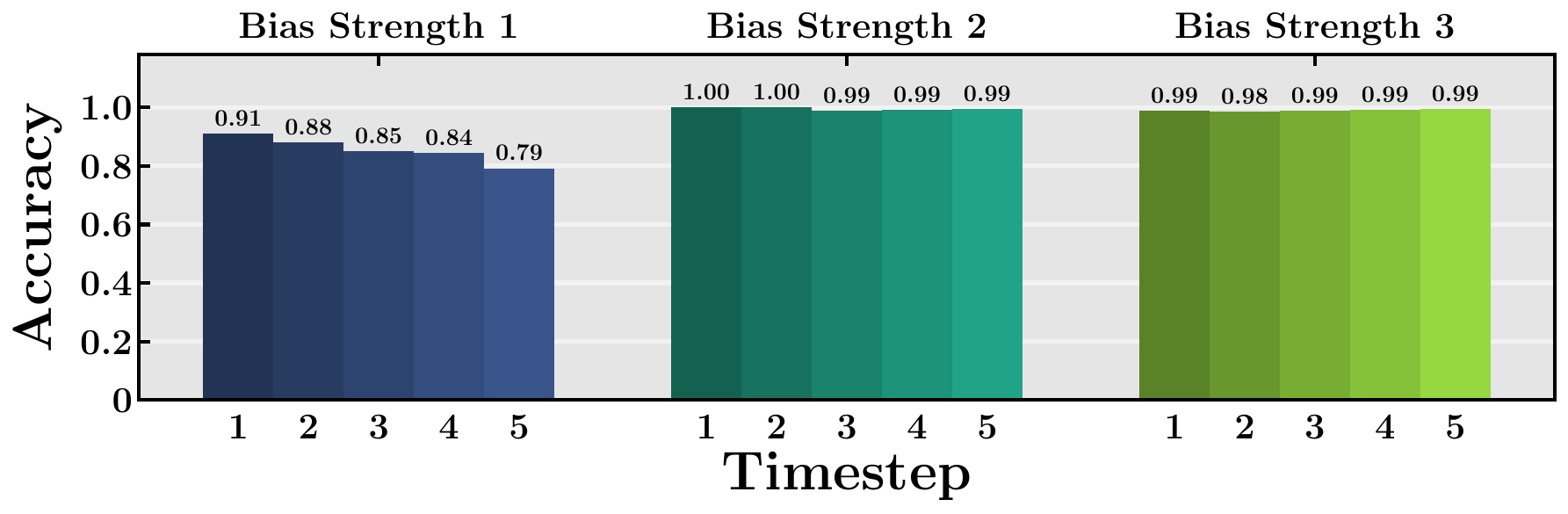}
    }
    \hfill
    \subfloat[Length 5 \& Population 3]{%
        \includegraphics[width=0.32\linewidth]{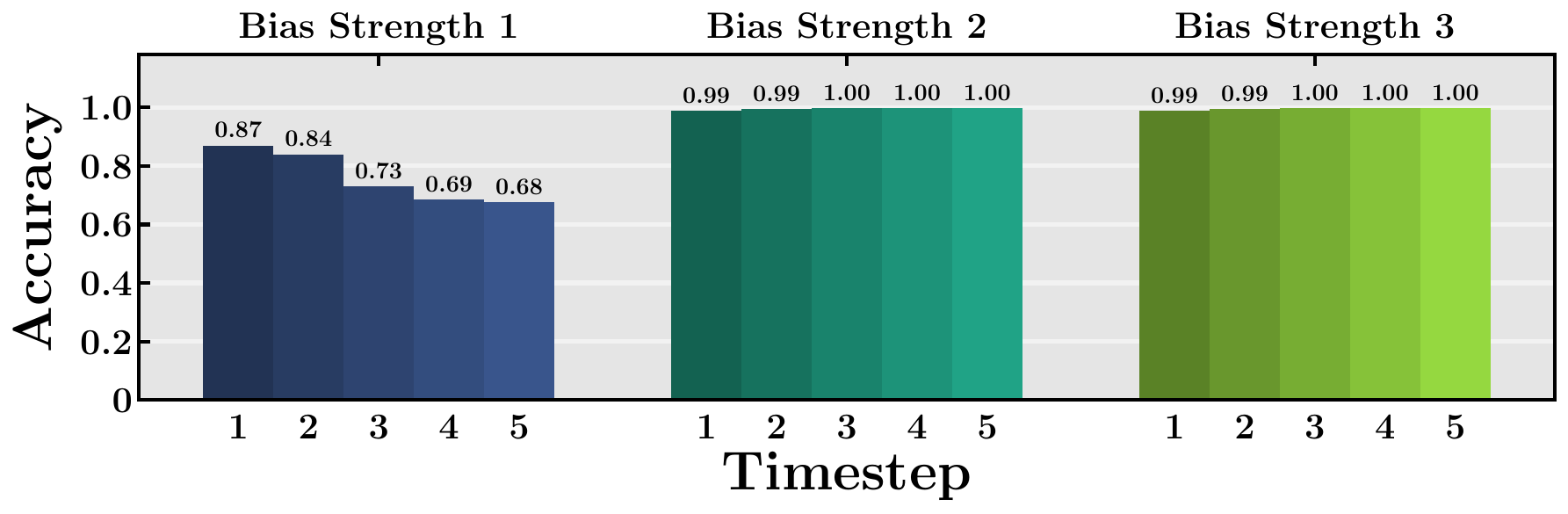}
    }
    \hfill
    \subfloat[Length 5 \& Population 4]{%
        \includegraphics[width=0.32\linewidth]{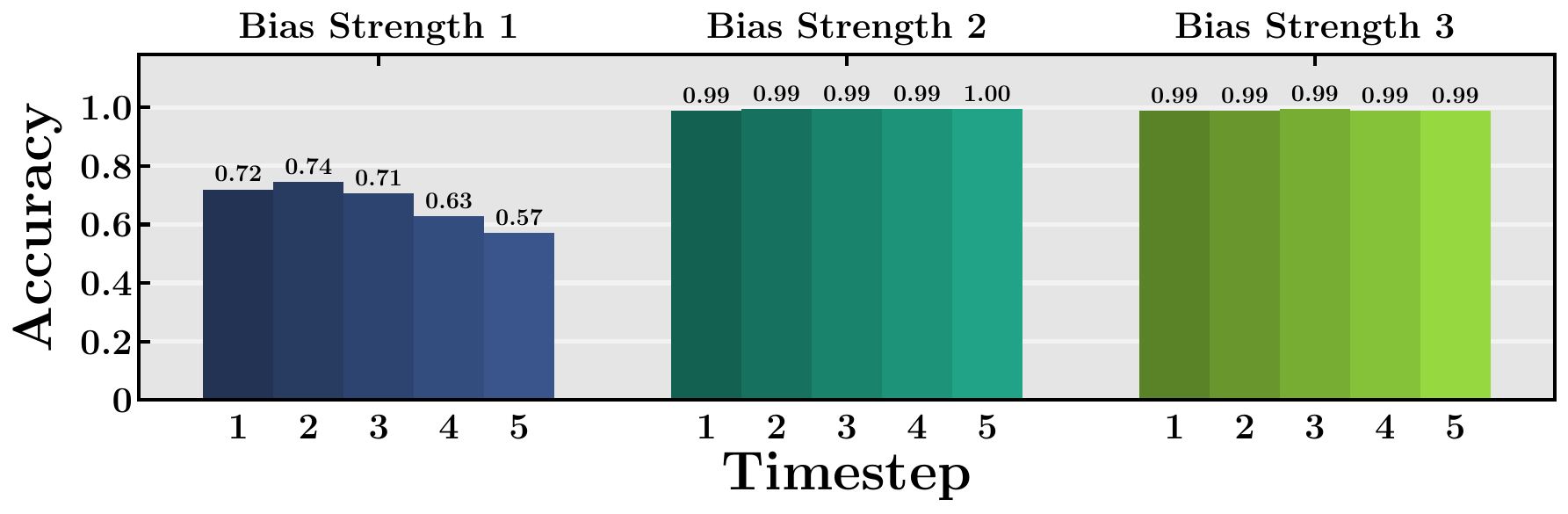}
    }
    
    \vfill

    \subfloat[Length 6 \& Population 2]{%
        \includegraphics[width=0.32\linewidth]{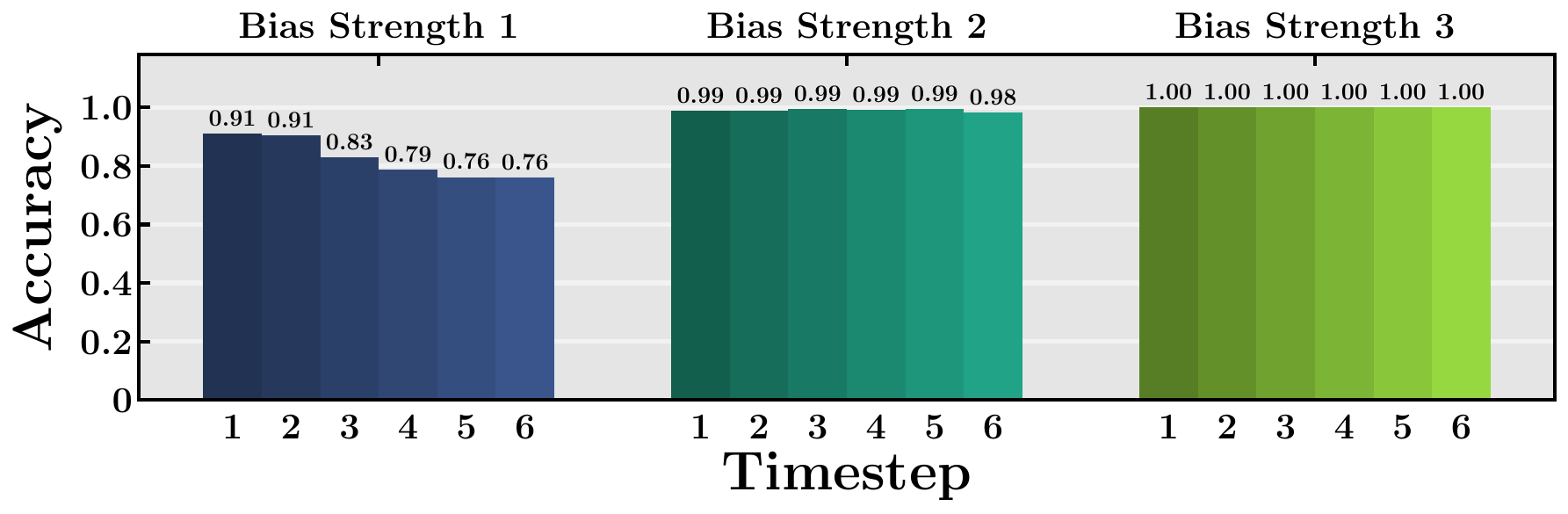}
    }
    \hfill
    \subfloat[Length 6 \& Population 3]{%
        \includegraphics[width=0.32\linewidth]{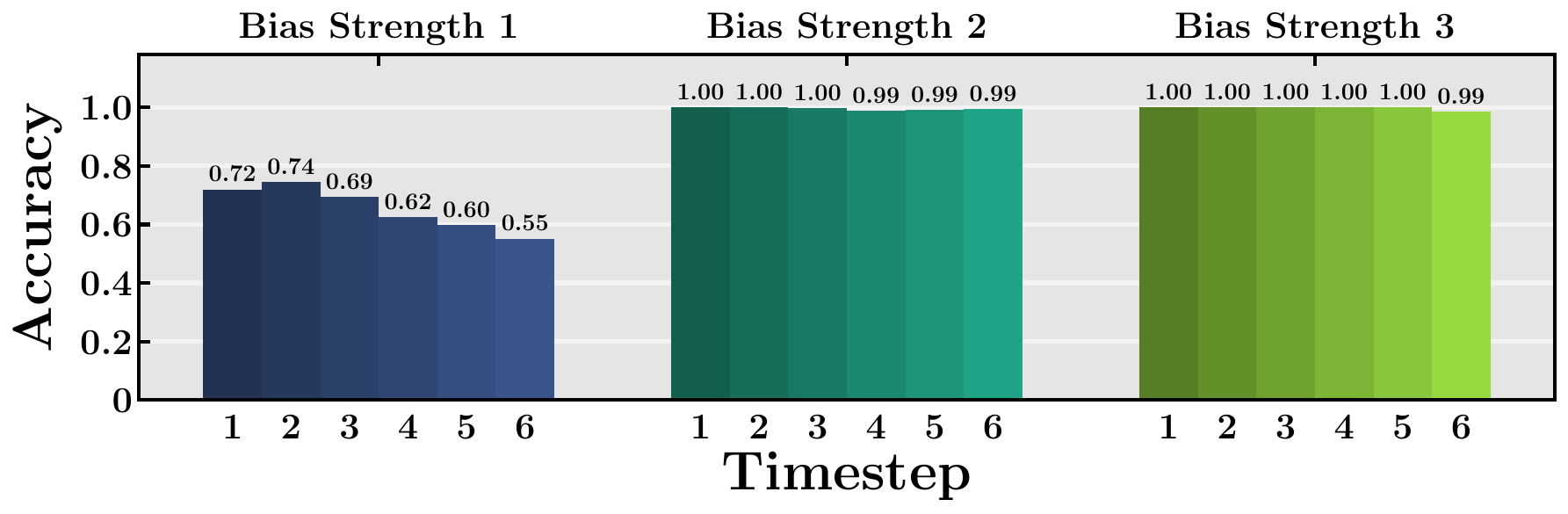}
    }
    \hfill
    \subfloat[Length 6 \& Population 4]{%
        \includegraphics[width=0.32\linewidth]{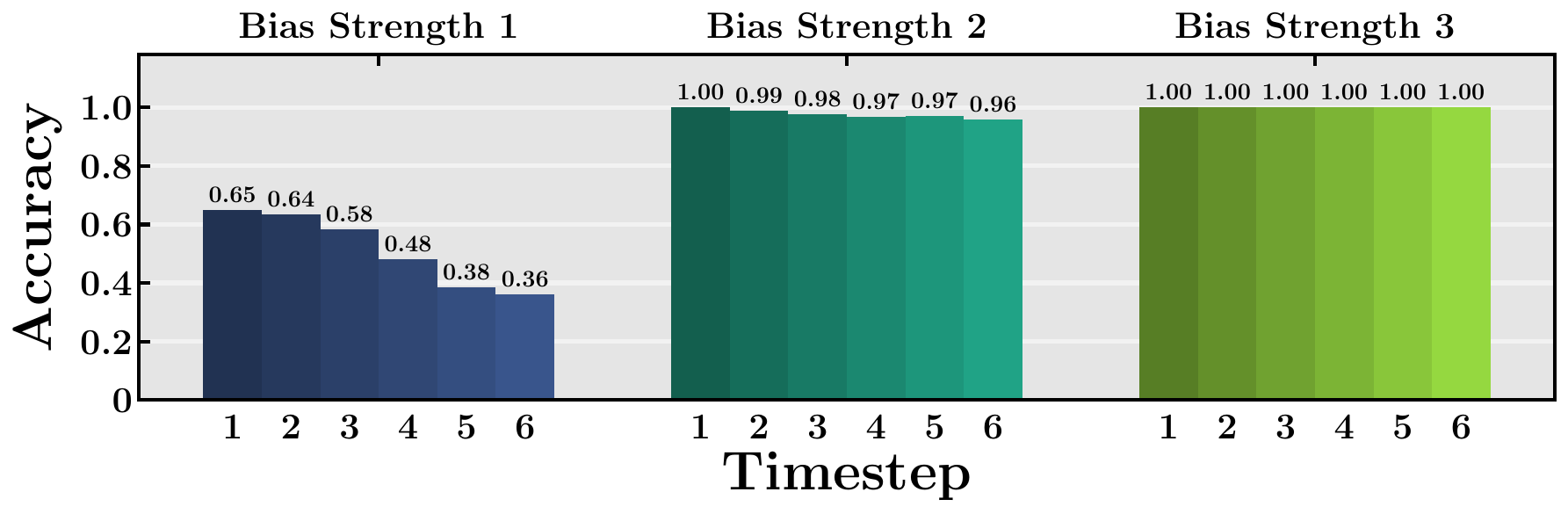}
    }

    \caption{Timestep-wise chronological accuracy under varying bias strengths, chronicle lengths and agent populations.}
    \label{fig:exp_acc_sub}
\end{figure*}

\begin{figure}[t!]
    \centering
    \subfloat[Length 6 \& Population 2]{%
        \includegraphics[width=0.99\linewidth]{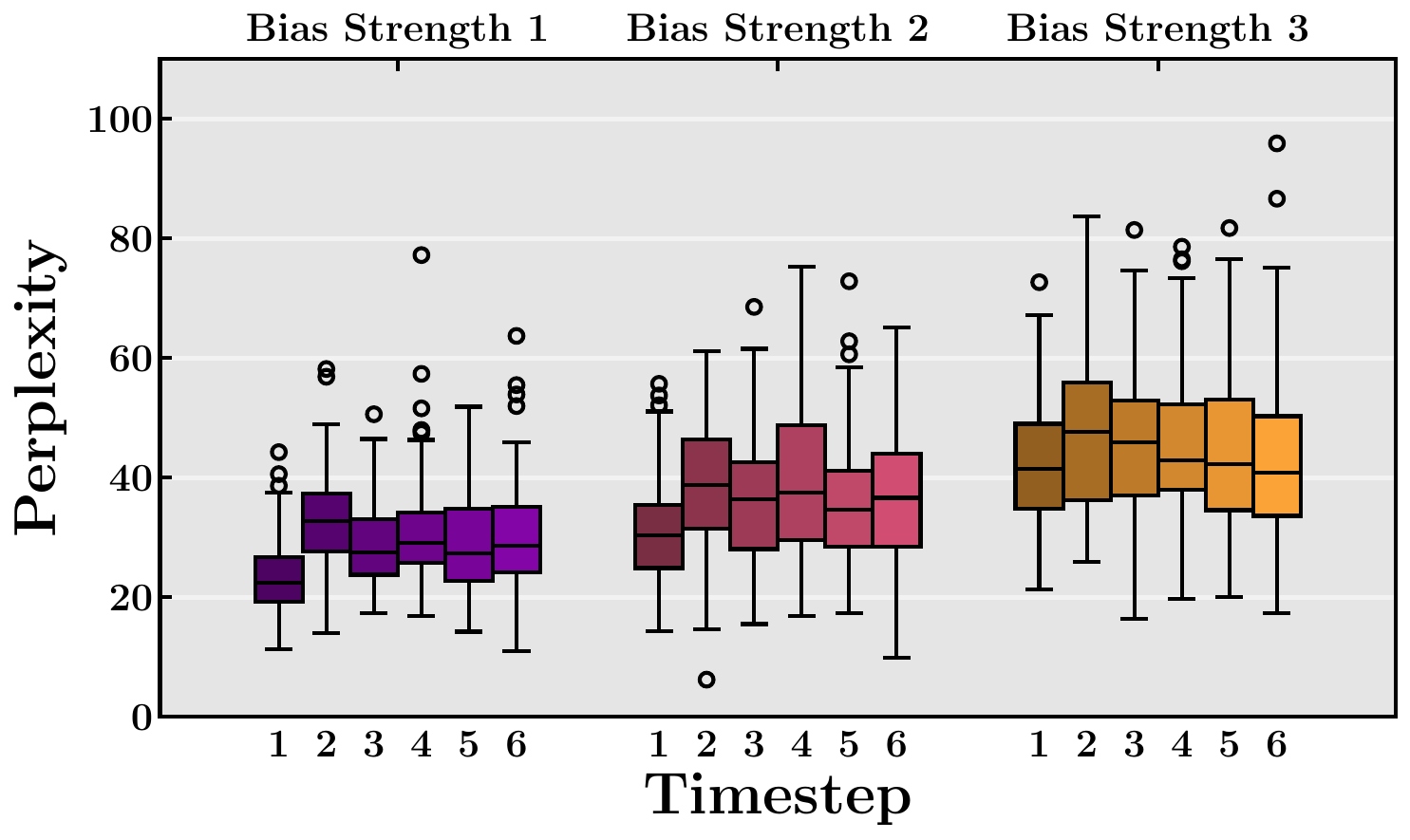}
    }
    \\
    \subfloat[Length 6 \& Population 3]{%
        \includegraphics[width=0.99\linewidth]{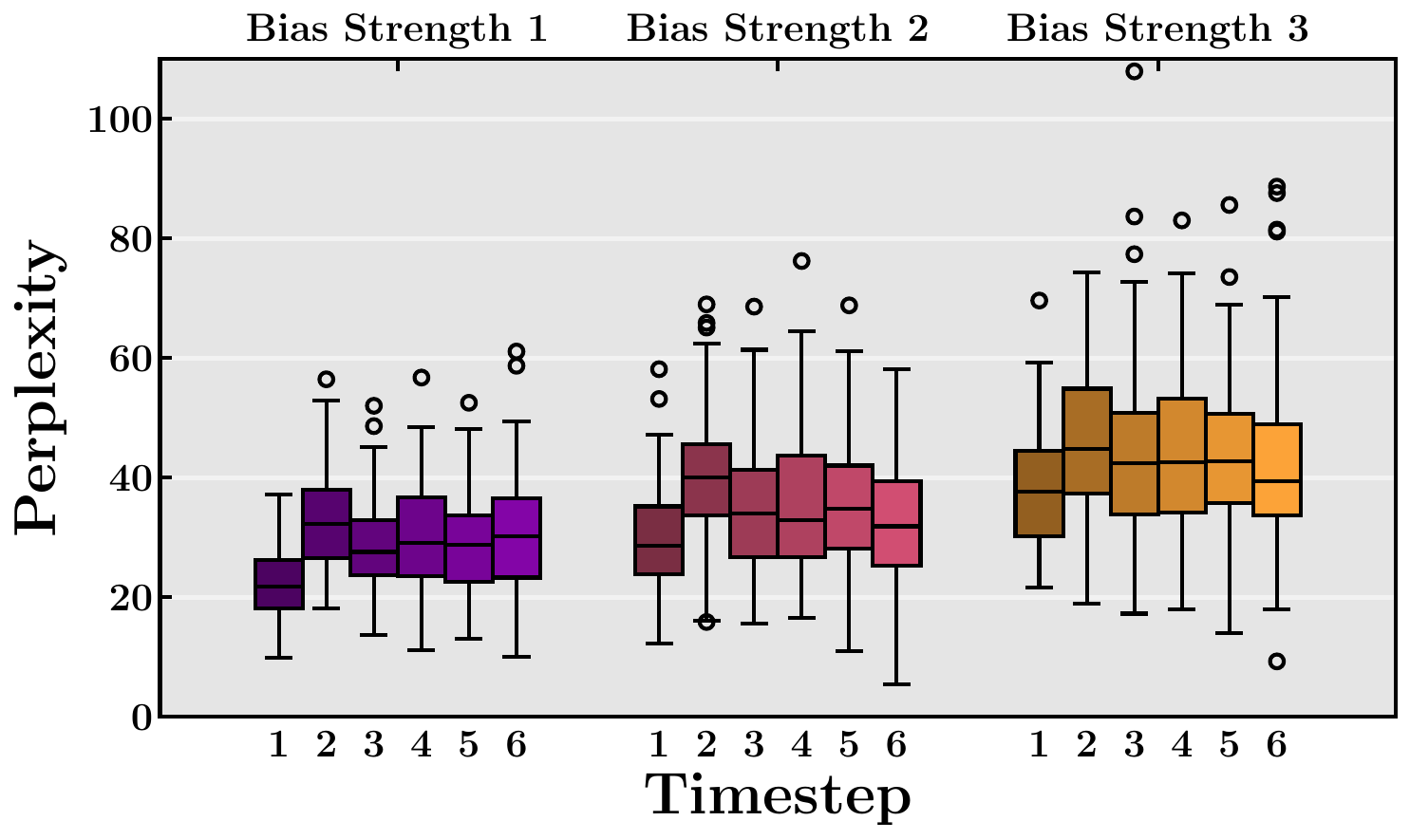}
    }
    \\
    \subfloat[Length 6 \& Population 4]{%
        \includegraphics[width=0.99\linewidth]{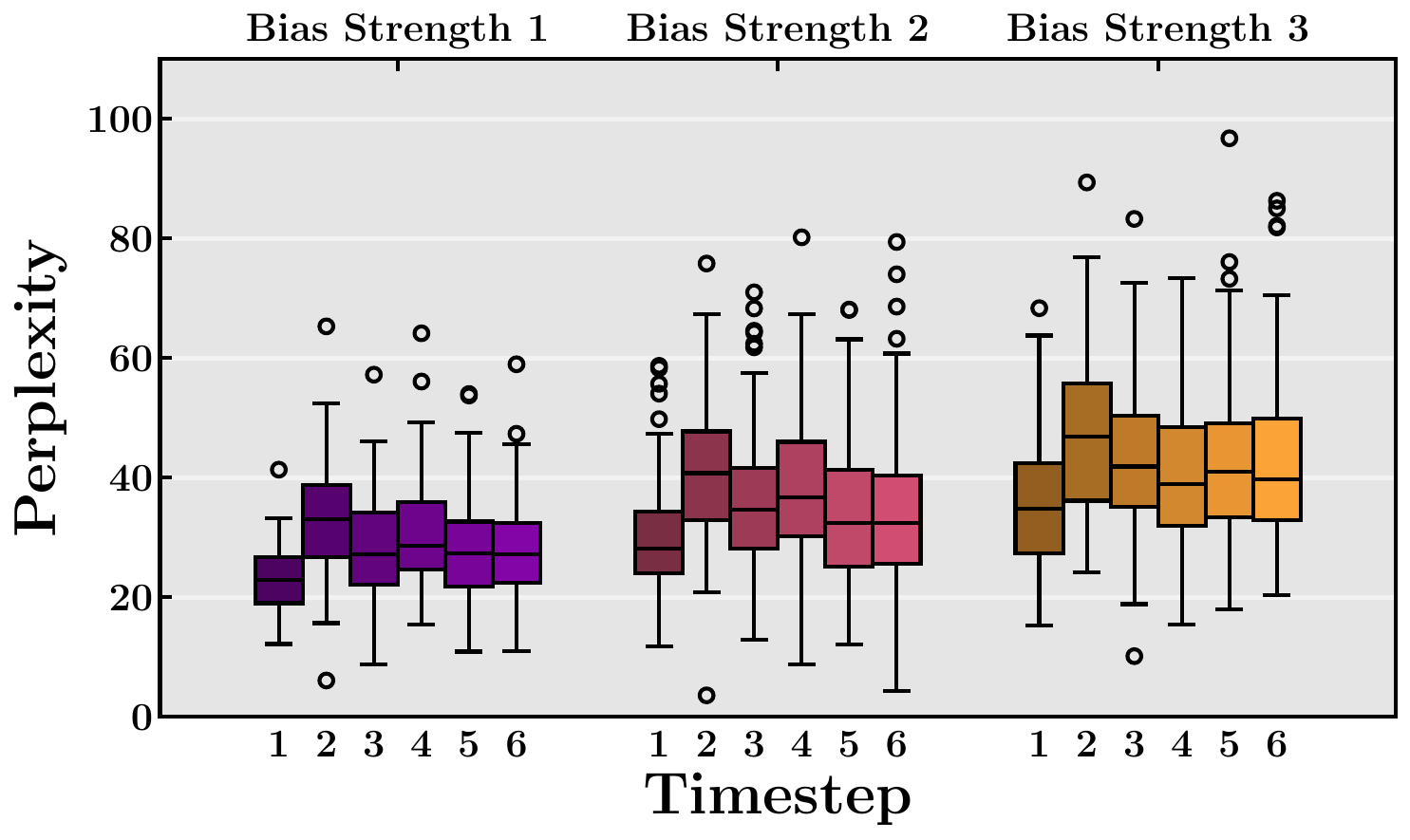}
    }
    
    \caption{Timestep-wise generative perplexity under varying bias strengths and agent populations with fixed chronicle length.}
    \label{fig:exp_ppl_sub}
\end{figure}

\begin{figure}[t!]
    \centering
    \subfloat[Length 6 \& Population 2]{%
        \includegraphics[width=0.99\linewidth]{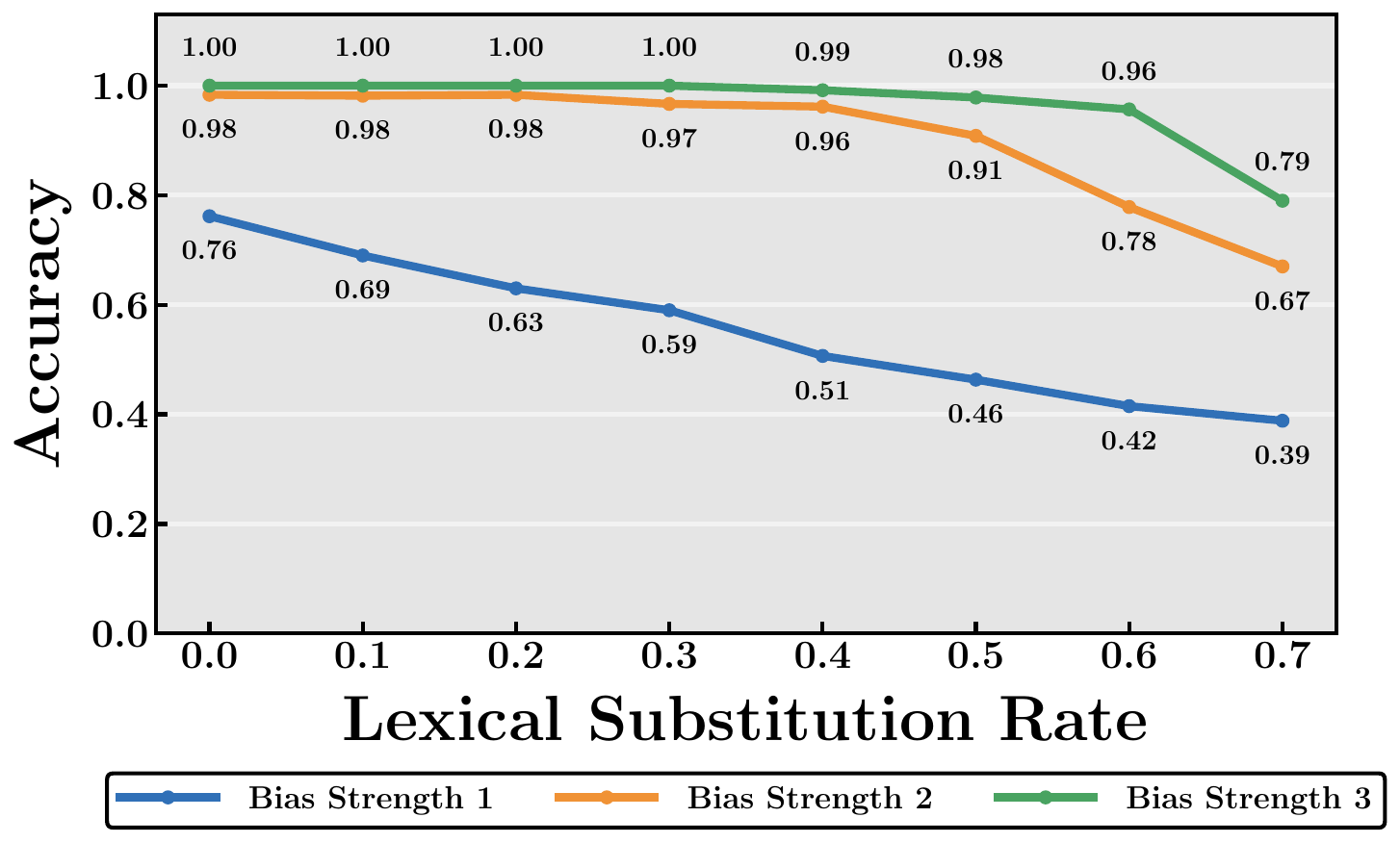}
    }
    \\
    \subfloat[Length 6 \& Population 3]{%
        \includegraphics[width=0.99\linewidth]{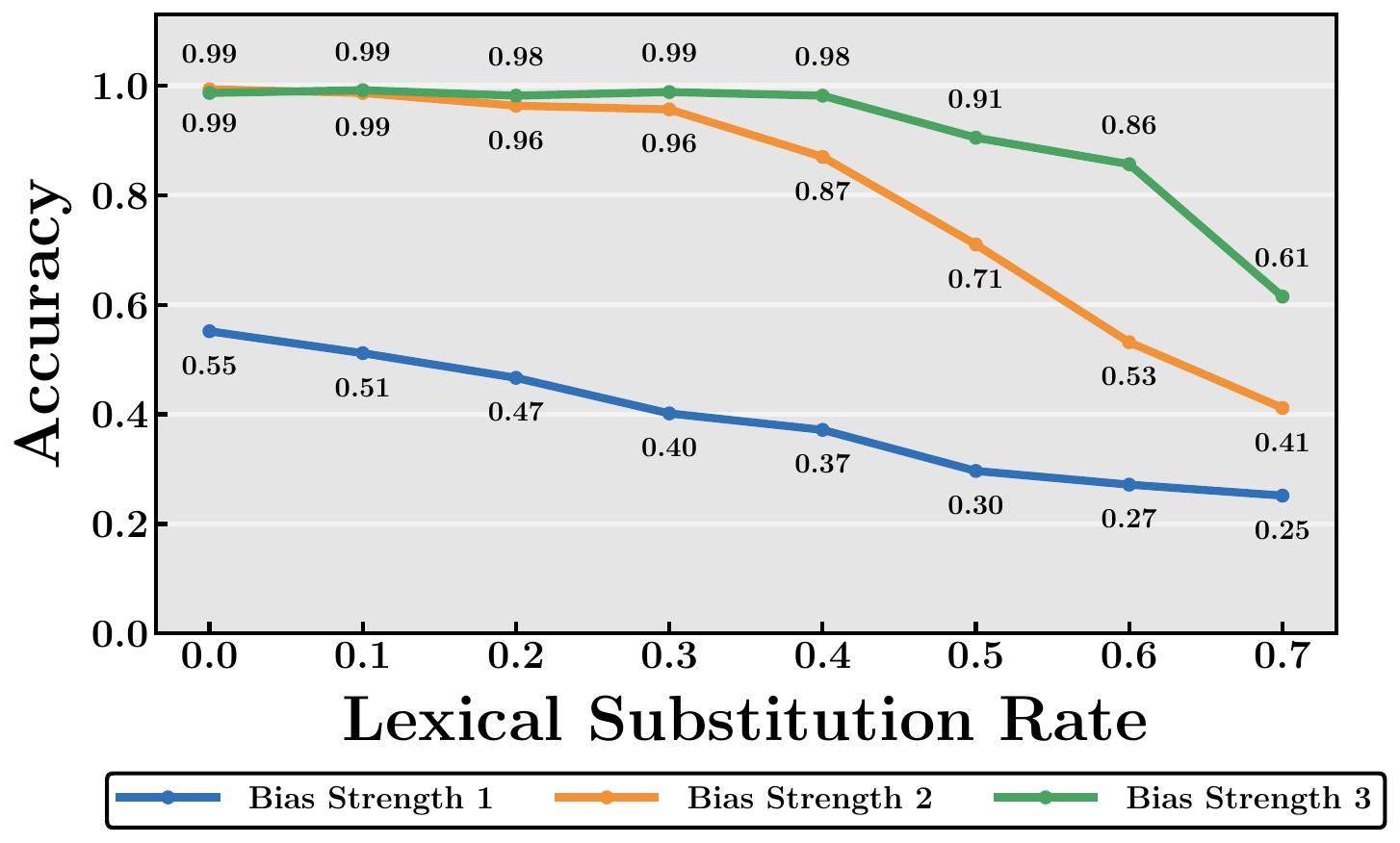}
    }
    \\
    \subfloat[Length 6 \& Population 4]{%
        \includegraphics[width=0.99\linewidth]{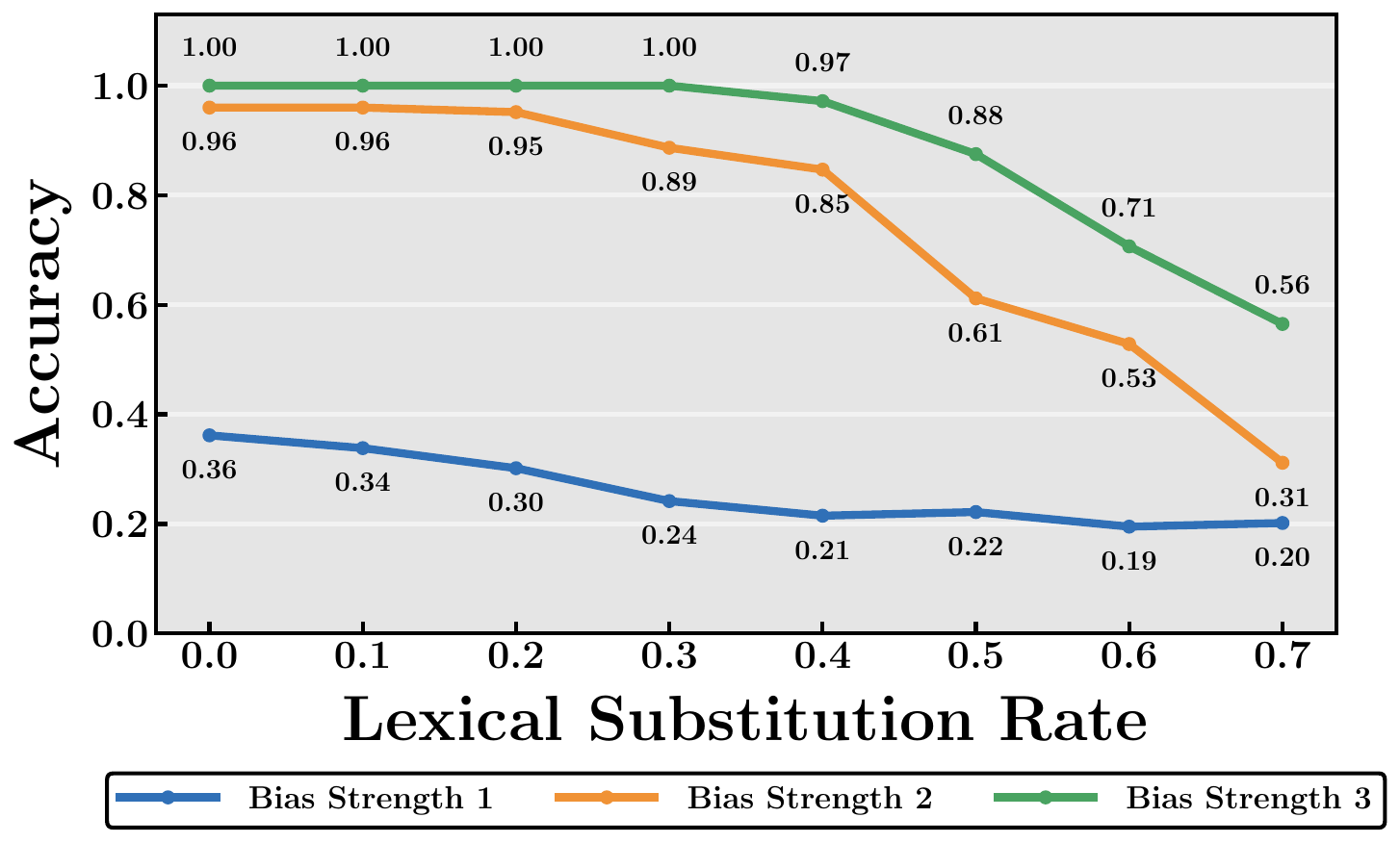}
    }
    
    \caption{Adversarial robustness against lexical substitution under varying bias strengths and agent populations with fixed chronicle length.}
    \label{fig:exp_rob_sub}
\end{figure}

\subsection{Chronicle Space Complexity}
Figure~\ref{fig:code_complexity} analyses the scaling effect of the number of valid chronicles $|\mathcal{X}|$ with respect to the chronicle length $T$ and the agent population $n$. Recall that the decoding process requires an exhaustive search over the entire set of valid chronicles to infer the most probable agent sequence from the generated text. As such, the number of valid chronicles directly determines the computational complexity of the decoding procedure. An exponential increase in the number of chronicles was evident as the chronicle length grows from 3 to 6 for varying numbers of agents from 2 to 4. This scaling effect reflects the combinatorial nature of multi-agent provenance tracking, where longer chronicles and larger agent populations lead to a combinatorial explosion in the size of the search space over possible chronicles.

\subsection{Chronological Accuracy}
Figure~\ref{fig:exp_acc_sub} examines the chronological performance using a timestep-wise accuracy metric. At each timestep $t$, accuracy (ACC) was computed as the proportion of correctly decoded chronicle symbols up to the current step, defined as:
\begin{equation} 
\text{ACC} = \frac{ \sum_{i=1}^{t} \mathbb{I}(\hat{x_i} = x_i) }{t} ,
\end{equation}
where $\hat{x}_i$ and $x_i$ denote the predicted and ground-truth symbols at timestep $i$, respectively, and $\mathbb{I}(\cdot)$ denotes the indicator function. This metric captures the phenomenon of error propagation, wherein an incorrectly decoded symbol is carried forward into subsequent chronicle updates and encoding steps, potentially leading to error accumulation. It was observed that the accuracy tended to decrease as the timestep increased, reflecting the challenge of preserving provenance integrity over extended generation horizons. Moreover, as the maximum chronicle length $T$ increased, the accuracy further declined due to the exponentially expanding size of the candidate chronicle space $|\mathcal{X}|$. This degradation, nevertheless, was alleviated by increasing the bias strength $\delta$. When the bias strength was raised from 1 to 2 or 3, near-perfect accuracy was consistently achieved across all experimental configurations, demonstrating the reliability of chronological identification under appropriate lexical bias.

\subsection{Generative Perplexity}
Figure~\ref{fig:exp_ppl_sub} evaluates the impact of increasing bias strength in the sampling process on generation quality. To quantify this effect, we measured the perplexity of the generated text, which captures the model's uncertainty in producing the observed sequence of tokens. Perplexity (PPL) represents the inverse of the average likelihood (i.e. the geometric mean) of the predicted token probabilities over a given sequence $\boldsymbol{v}$, or equivalently, the exponentiated average negative log-likelihood, defined as:
\begin{equation}
\begin{split}
	\text{PPL} =\left(\prod _{i=1}^{ |\boldsymbol{v}| } p_{v_i} \right)^{-\frac{1}{ | \boldsymbol{v} |}} = \exp \left( -\frac{1}{ | \boldsymbol{v} | } \sum_{i=1}^{ |\boldsymbol{v}| } \log p_{v_i} \right) ,
\end{split}
\end{equation}
where $p_{v_i}$ denotes the probability assigned by the model to the $i$-th token in the sequence $\boldsymbol{v}$, and $|\boldsymbol{v}|$ is the length of the sequence. Higher perplexity indicates that the model assigns lower confidence to the generated tokens, reflecting increased uncertainty or degradation in generation quality. For this analysis, the chronicle length was fixed at 6, and perplexity was computed at each timestep under varying bias strengths and numbers of agents. As expected, increasing the bias strength led to higher perplexity, indicating a degradation in generation fluency and naturalness. While the first generation step typically exhibited lower perplexity, subsequent steps did not show a consistent increasing trend, suggesting that the impact of biased sampling stabilises after the initial generation step. Moreover, the number of agents involved did not appear to have a substantial impact on perplexity, which might imply that generation quality was primarily influenced by the bias strength rather than the complexity of the multi-agent setting.

\subsection{Adversarial Robustness}
Figure~\ref{fig:exp_rob_sub} assesses the system's robustness against adversarial attacks intended to obscure the provenance of agent interactions and thereby undermine accountability in the AI ecosystem. A basic adversarial attack is lexical substitution, in which words in the generated text are replaced with semantically equivalent alternatives (typically synonyms). To simulate lexical substitution, we replaced a controlled fraction of words (including nouns, verbs, adjectives and adverbs) with random synonyms drawn from WordNet synsets, increasing the substitution rate from 0\% to 70\% in 10\% increments. We measured chronological accuracy for a chronicle length of 6 under varying agent populations and bias strengths. Under low bias strength ($\delta = 1$), accuracy degraded steadily as the substitution rate increased, eventually approaching the random-guess baseline of $1/(n+1)$, corresponding to uniform guessing over the symbol space of $n$ agents plus an unassigned agent. This degradation arises from the disruption of alignment between the generated tokens and the codeword-defined vocabulary subset, which dilutes the statistical signal for chronicle reconstruction. Under high bias strength ($\delta = 3$), we observed a significant improvement in adversarial robustness: for population 2, accuracy remained above 90\% even when up to 60\% of words were substituted; for population 3, this threshold fell between 50\% and 60\%; and for population 4, between 40\% and 50\%. These trends confirm that a stronger sampling bias can, to some extent, counteract aggressive lexical substitution by concentrating the sampling distribution on codeword-associated tokens, thereby forming more robust statistical features.

\section{Discussion}
This section reflects on broader aspects of the proposed chronological system. We examine the assumptions underlying the experimental design and the implications of the empirical results, while outlining open directions for future research. The discussion aims to bridge the gap between the theoretical formulation and practical deployment.

\subsection{Agent Specialisation}
In this study, all agents are instantiated from a single underlying language model and assigned the same continual writing task, resulting in a minimal and controlled experimental environment. The use of homogeneous agents allows for the isolation of the proposed chronological system from confounding factors associated with application-dependent model variability. The task formulation intentionally discards previous generative traces, approximating a scenario in which no explicit memory of prior contributions is preserved. A practical constraint arises from the requirement that all agents share a common vocabulary space, as the codebook for encoding and decoding is defined over this space. In contrast, real-world deployments are expected to involve specialised agents with application-specific characteristics and capabilities. Extending the methodology to heterogeneous agents therefore constitutes an open direction for future investigation.

\subsection{Interaction Dynamics}
A fundamental limitation on scalability arises from the combinatorial explosion of the chronicle space. As the number of participating agents or the length of interactions increases, the space of possible chronicles expands exponentially, rendering computational costs impractical. To mitigate this limitation, the structure of interaction dynamics may be exploited. In practical applications, agent interactions may not be uniform or unconstrained. Certain agents may be more likely to follow specific predecessors, or be restricted to operate within particular stages of a task. Such non-uniform interaction patterns induce a structural prior over the chronicle space, effectively reducing the set of plausible sequences. Incorporating these application-specific priors into the methodological design may enable adaptive reduction of the chronicle space, yielding a more scalable and context-aware framework.

\subsection{Parametric Trade-off}
The experimental results reveal an inherent trade-off between the reliability of chronological attribution and the quality of text generation, governed by the bias strength parameter. In essence, a stronger bias restricts the vocabulary available during the generation process. Increasing the bias strength amplifies the distinguishability of statistical patterns, thereby improving the accuracy of chronological attribution. However, this comes at the cost of constraining the token sampling process, leading to higher perplexity and reduced linguistic fluency. This parametric trade-off reflects a fundamental tension between attribution accuracy and generative quality, positioning the bias strength as a tunable variable for application-specific requirements.

\subsection{Noisy Communication}
The robustness evaluation in this study focuses on lexical substitution, wherein words are replaced with semantically equivalent alternatives. This form of linguistic perturbation can be interpreted as noise in a communication channel. In practical scenarios, a broader range of noisy channel models may arise. Beyond lexical substitution, perturbations that preserve semantics include syntactic restructuring, paraphrasing, rewriting, or translation across languages. The generated text may also be subject to channel noise such as deletion or truncation, where portions of the text are erased and lost. Conversely, in certain controlled and secure environments, the content may remain largely unaltered throughout interactions. These diverse conditions suggest that robustness is inherently shaped by the characteristics of the underlying communication channel. This calls for the formulation of application-specific noise models and the adaptations of the proposed methodology to suit individual deployment scenarios.

\section{Conclusion}

The problem of tracking multi-agent provenance amidst the continual act of shared creation was investigated in this study. We introduced a chronological system for tracking provenance in language generation, where the history of agent contributions is not explicitly recorded as metadata but embedded within the generated content itself through the process of sampling lexical tokens. Experimental results validated the performance of the proposed system under conditions of sequential content overwriting. Nevertheless, the evaluation also revealed several dimensions that warrant further research. To begin with, the combinatorial growth governed by chronicle length and agent population reflects scaling challenges, suggesting the need for efficient systems that support longer or variable-length chronicles. Realistic deployment often involves uncertain interaction lengths that may not be known in advance, making the assumption of a fixed chronicle length a practical limitation. In addition, while raising the lexical bias strength substantially improves chronological accuracy, it also elevates linguistic perplexity, exposing an inherent trade-off between forensic traceability and generation quality. Furthermore, under more adaptive adversarial conditions beyond lexical perturbations, further security measures and regulatory frameworks may be required to maintain provenance integrity. Towards broader applications, future research may explore chronological provenance in more complex cyber ecosystems, including multimodal integration and human-in-the-loop generation, where provenance traceability constitutes a foundational tenet for trustworthy AI.

\section*{Acknowledgement}
{This work was supported in part by the Japan Society for the Promotion of Science (JSPS) under KAKENHI Grants JP21H04907 and JP24H00732, and in part by the Japan Science and Technology Agency (JST) under the CREST Grants JPMJCR20D3 and JPMJCR2562, including the AIP Challenge Program, and under the AIP Acceleration Grant JPMJCR24U3 and under the K Program Grant JPMJKP24C2.}

\bibliography{Transactions-Bibliography/bstcontrol, Bib/bib_chronicle}
\bibliographystyle{Transactions-Bibliography/IEEEtran}

\vspace{-12pt}
\begin{IEEEbiographynophoto}{Ching-Chun Chang} received the PhD in Computer Science from the University of Warwick, UK, in 2019. He is currently affiliated with the National Institute of Informatics, Japan, as a Project Assistant Professor. He also serves as a Visiting Researcher at Peking University, China, and a Distinguished Professor at Hangzhou Dianzi University, China. He participated in the Short-Term Scientific Mission supported by European Cooperation in Science and Technology Actions at the Faculty of Computer Science, Otto von Guericke University of Magdeburg, Germany, in 2016. He was granted the Marie-Curie Fellowship and participated in the Research and Innovation Staff Exchange supported by Marie Skłodowska-Curie Actions at the Department of Electrical and Computer Engineering, New Jersey Institute of Technology, USA, in 2017. He was a Visiting Scholar at the School of Computing and Mathematics, Charles Sturt University, Australia, in 2018, and at the School of Information Technology, Deakin University, Australia, in 2019. He was a Research Fellow at the Department of Electronic Engineering, Tsinghua University, China, in 2020. His research interests include artificial intelligence, biometrics, cryptography, cybersecurity, evolutionary computation, forensics, information theory, steganography, and watermarking.
\end{IEEEbiographynophoto}

\vspace{-2pt}
\begin{IEEEbiographynophoto}{Isao Echizen} received BS, MS, and DE degrees from the Tokyo Institute of Technology, Japan, in 1995, 1997 and 2003, respectively. He joined Hitachi, Ltd. in 1997 and until 2007 was a Research Engineer in the company's systems development laboratory. He is currently a Director and Professor of the Information and Society Research Division, as well as a Director of the Global Research Center for Synthetic Media, at the National Institute of Informatics; a Professor in the Department of Information and Communication Engineering, Graduate School of Information Science and Technology, the University of Tokyo; and a Professor in the Graduate University for Advanced Studies (SOKENDAI), Japan. He was a Visiting Professor at the Tsuda University, Japan; at the University of Freiburg, Germany; and at the University of Halle-Wittenberg, Germany. He is currently engaged in research on AI security, multimedia security and multimedia forensics, serving as a Research Director for the CREST FakeMedia project and the K Program SYNTHETIQ X project of the Japan Science and Technology Agency (JST). He received the Commendation for Science and Technology by the Minister of Education, Culture, Sports, Science and Technology (Research Category) in 2025. He also received the IEICE Best Paper Award in 2023; the IPSJ Best Paper Awards in 2005 and 2014; the IPSJ Nagao Special Researcher Award in 2011; the DOCOMO Mobile Science Award in 2014; the IISEC Information Security Cultural Award in 2016; and the IEEE WIFS Best Paper Award in 2017. He is an IEICE Fellow, an IPSJ Fellow, an IEEE Senior Member, an IFIP Japanese Representative, and an APSIPA Vice President.
\end{IEEEbiographynophoto}

\end{document}